\documentclass[11pt]{article}

\usepackage{wrapfig}
\usepackage{graphicx}
\usepackage{natbib}

\usepackage{booktabs}
\usepackage{subcaption}
\usepackage{cprotect}
\usepackage{amssymb,amsmath,amsthm,mathtools}
\usepackage{epstopdf}
\usepackage{bbm}
\usepackage[marginratio=1:1,height=600pt,width=460pt,tmargin=100pt]{geometry}
\usepackage{layout, color}
\usepackage{bbm}
\usepackage{enumerate}
\usepackage{authblk}

\usepackage{algorithmic}
\usepackage{algorithm}

\usepackage{setspace}
\usepackage{hyperref}

\usepackage{url}

\usepackage{caption}
\usepackage{subcaption}

\usepackage{mdwlist}
\usepackage{multirow}
\usepackage{amsthm}
\usepackage{color}
\usepackage{makecell}

\theoremstyle{remark}
\newtheorem{remark}{Remark}

\newtheorem{conjecture*}{Conjecture}
\theoremstyle{plain}

\newlength{\tempdima}
\newcommand{\rowname}[1]
{\rotatebox{90}{\makebox[\tempdima][c]{\textbf{#1}}}}
\renewcommand{\thesubfigure}{\alph{subfigure}}
\newcommand{\mycaption}[1]
{\refstepcounter{subfigure}\textbf{(\thesubfigure) }{\ignorespaces #1}}

\usepackage{array}
\newcolumntype{P}[1]{>{\centering\arraybackslash}p{#1}}

\newcommand{\JKO}{JKO-iFlow}

\newcommand{\flowname}{\text{Q-flow }}
\newcommand{\rationame}{\text{flow-ratio }}

\newcommand{\ours}{\text{Ours}}
\newcommand{\DREinf}{DRE-$\infty$}

\newcommand{\R}{\mathbb{R}}
\newcommand{\E}{\mathbb{E}}
\newcommand{\calN}{\mathcal{N}}

\newcommand{\calL}{\mathcal{L}}

\usepackage{soul}
\usepackage{xcolor}

\begin{document}

\title{Computing high-dimensional optimal transport \\ by flow neural networks}

\author[1]{Chen Xu}
\author[2]{Xiuyuan Cheng}
\author[1]{Yao Xie}
\affil[1]{{\small H. Milton Stewart School of Industrial and Systems Engineering, Georgia Institute of Technology.}}
\affil[2]{{\small Department of Mathematics, Duke University}}

\date{\vspace{-30pt}}

\maketitle

\begin{abstract}
  Computing optimal transport (OT) for general high-dimensional data has been a long-standing challenge. 
Despite much progress, most of the efforts including neural network methods have been focused on the static formulation of the OT problem. 
The current work proposes to compute the dynamic OT between two arbitrary distributions $P$ and $Q$ by optimizing a flow model, where both distributions are only accessible via finite samples.
Our method learns the dynamic OT by finding an invertible flow that minimizes the transport cost.  
The trained optimal transport flow subsequently allows for performing many downstream tasks, including infinitesimal density ratio estimation (DRE) and
domain adaptation by interpolating distributions in the latent space. The effectiveness of the proposed model on high-dimensional data is demonstrated by strong empirical performance on OT baselines,
image-to-image translation, 
and high-dimensional DRE.
\end{abstract}

\section{Introduction}

The problem of finding a transport map between two general distributions $P$ and $Q$ in high dimension is essential in statistics, optimization, and machine learning. 
When both distributions are only accessible via finite samples, the transport map needs to be learned from the data. Despite the modeling and computational challenges, this setting has applications in many fields. 
For example, transfer learning in domain adaption aims to obtain a model on the target domain at a lower cost
by making use of an existing pre-trained model on the source domain \citep{courty2014domain,courty2017joint}, 
and this can be achieved by transporting the source domain samples to the target domain using the transport map. 
The (optimal) transport has also been applied to achieve model fairness \citep{silvia2020general}.
By transporting distributions corresponding to different sensitive attributes to a common distribution, an unfair model is calibrated to match certain desired fairness criteria (e.g., demographic parity \citep{jiang2020wasserstein}). 
Additionally, the transport map can provide intermediate interpolating distributions between $P$ and $Q$.
In density ratio estimation, this bridging facilitates the so-called ``telescopic'' DRE \citep{rhodes2020telescoping}, which has been shown to be more accurate when $P$ and $Q$ significantly differ. Furthermore, learning such a transport map between two sets of images can help solve problems in computer vision, such as image-to-image translation \citep{isola2017image}.

This work focuses on a continuous-time formulation of the problem where we are to find an invertible transport map  $T_t:\mathbb R^d\rightarrow \mathbb R^d$ continuously parameterized by time $t \in [0,1]$ and satisfying that $T_0 = {\rm Id}$ (the identity map) and $(T_1)_\# P = Q$. Here we denote by $T_\# P $ the push-forward of distribution $P$ by a mapping $T$, such that $(T_\# P)(\cdot) = P( T^{-1} (\cdot))$.
Suppose $P$ and $Q$ have densities $p$ and $q$ respectively in $\R^d$ (we also use the push-forward notation $_\#$ on densities), the transport map $T_t$ defines 
\[
\rho(x,t) := (T_t)_\# p, \quad  \,  \text{s.t.}  \quad  \rho(x,0) = p, \quad \rho(x,1)=q.
\]
We will adopt the neural Ordinary Differential Equation (ODE) approach where we represent $T_t$ as the solution map of an ODE, 
{whose velocity field is parameterized by a neural network \citep{chen2018neural}.}
The resulting map $T_t$ is invertible, and the inversion can be computed by integrating the neural ODE reverse in time. 
Our model learns the flow from two sets of finite samples from $P$ and $Q$. 
The velocity field is optimized to minimize the transport cost to approximate the optimal velocity in the dynamic OT formulation, i.e., the Benamou-Brenier equation. 

The neural-ODE model has been intensively developed in Continuous Normalizing Flows (CNF) \citep{nflow_review}.
In CNF, the continuous-time flow model, usually parameterized by a neural ODE, transports from a data distribution $P$ (accessible via finite samples) to a terminal analytical distribution, which is typically the normal one $\calN(0, I_d)$, per the name ``normalizing''.
The study of normalizing flow dated back to non-deep models with statistical applications \citep{tabak2010density}, and deep CNFs have recently developed into a popular tool for generative models and likelihood inference of high dimensional data. 
CNF models rely on the analytical expression of the terminal distribution in training. 
Since our model is also a flow model that transports from data distribution $P$ to a general (unknown) data distribution $Q$, both accessible via empirical samples, we name our model ``Q-flow'' which is inspired by the CNF literature. 
{We emphasize that our motivation is to solve high dimensional OT by a flow model rather than CNF generative applications.}

After developing a general approach to the \flowname model, in the second half of the paper, we focus on the application of telescopic DRE.
After training a \flowname model (the ``flow net''),
we leverage the intermediate densities $\rho(x,t)$, 
which is accessed by finite sampled by pushing the $P$ samples by $T_t$,
to train an additional continuous-time classification network (the ``ratio net'') over time $t \in [0,1]$.
The ratio net estimates the infinitesimal change of the log-density $\log \rho(x,t)$ over time,
and its time-integral from 0 to 1 yields an estimate of the (log) ratio $q/p$. 
The efficiency of the proposed \flowname net and the infinitesimal DRE net is experimentally demonstrated on high-dimensional data. 

In summary, the contributions of the work include:


  \raisebox{0.25ex}{\tiny$\bullet$}  We develop a flow-based model {\it\flowname}net to learn a continuous invertible transport map between arbitrary pair of distributions $P$ and $Q$ in $\mathbb{R}^d$ from two sets of data samples.
    We propose training a neural ODE model to minimize the transport cost so that the flow approximates the optimal transport in dynamic OT.
    The end-to-end training of the model refines any initial flow that may not attain the optimal transport, e.g., obtained by training two CNFs or other interpolating schemes.

    %
    \raisebox{0.25ex}{\tiny$\bullet$} Leveraging a trained \flowname net, we propose to train a separate continuous-time network,
    called {\it \rationame}net,
    to perform infinitesimal DRE from $P$ to $Q$ given finite samples.
    The \rationame net is trained by minimizing a classification loss to distinguish neighboring distributions on a discrete-time grid along the flow, and it improves the performance over prior models on high-dimensional mutual information estimation and energy-based generative models.

    %
   \raisebox{0.25ex}{\tiny$\bullet$} We show the effectiveness of the \flowname net on simulated and real data.
    On public OT benchmarks, %
    we demonstrate improved performance over popular baselines.
    On the image-to-image translation task, 
    the proposed \flowname learns a trajectory from 
    an input image to a target sample that resembles the input in style, 
    and it also achieves comparable or better quantitative metrics than the state-of-the-art neural OT model.

\subsection{Related Works}

\paragraph{Normalizing flows.} 
When the target distribution $Q$ is an isotropic Gaussian $\calN(0, I_d)$, normalizing flow models have demonstrated vast empirical successes in building an invertible transport $T_t$ between $P$ and $\calN(0, I_d)$ \citep{nflow_review,papamakarios2021normalizing}. The transport is parameterized by deep neural networks, whose parameters are trained via minimizing the Kullback-Leibler (KL)-divergence between transported distribution $(T_1)_{\#} P $ and $\calN(0, I_d)$. Various continuous \citep{FFJORD,finlay2020train} and discrete \citep{dinh2017density,iResnet} normalizing flow models have been developed, along with proposed regularization techniques \citep{onken2021otflow,xu2022iGNN,xu2023normalizing} that facilitate the training of such models in practice. 
Since our \flowname can be viewed as a transport-regularized flow between $P$ and $Q$, we further review related works on building normalizing flow models with transport regularization. 
\citep{finlay2020train} trained the flow trajectory with regularization based on $\ell_2$ transport cost and Jacobian norm of the network-parameterized velocity field. 
\citep{onken2021otflow} proposed to regularize the flow trajectory by $\ell_2$ transport cost and the deviation from the Hamilton–Jacobi–Bellman (HJB) equation. These regularizations have been shown to improve effectively 
over un-regularized models at a reduced computational cost. 
Regularized normalizing flow models have also been used to solve high dimensional Fokker-Planck equations \citep{liu2022neural} and mean-field games \citep{huang2023bridging}.

\paragraph{Distribution interpolation by neural networks.}
Several works have been done to establish a continuous-time interpolation between general high-dimensional distributions.
\citep{albergo2023building} proposed to use a stochastic interpolant map between two arbitrary distributions and train a neural network parameterized velocity field to transport the distribution along the interpolated trajectory, {and the method is also known as Flow Matching}. 
\citep{neklyudov2023action} proposed an action matching scheme that leverages a pre-specified trajectory between $P$ and $Q$ to learn the OT map between two \textit{infinitesimally close} distributions along the trajectory.
Similar to Flow-Matching methods \citep{albergo2023building,lipman2023flow,liu2022rectified}, our approach also computes a deterministic probability transport map. %
However, the interpolant mapping used in these prior works is generally not the optimal transport interpolation,
while our proposed \flowname is optimized to find the optimal velocity in dynamic OT (see Section \ref{sec_prelim}). 
Generally, the flow attaining optimal transport can improve model efficiency and generalization performance \citep{huang2023bridging}, and in this work we experimentally show the benefits in high-dimensional DRE and image-to-image translation.

\paragraph{Computation of OT.}

Many mathematical theories and computational tools have been developed to tackle the OT problem
\citep{villani2009optimal,benamou2000computational,peyre2019computational}.
In this work, we focus on the Wasserstein-2 distance, which suffices for many applications. 
Compared to non-deep approaches,
neural network OT methods enjoy scalability to high dimensional data;
However, most works in the literature adopt the 
\textit{static} OT formulation
\citep{xie2019scalable,huang2021convex,morel2023turning,fan2023neural,korotin2021wasserstein,korotin2023neural,amos2023on}.
By static OT, we mean the problem that, in Monge formulation, looks for a transport $T$ that minimizes $\mathbb{E}_{x\sim P}\|x-T(x)\|^2 $ and satisfies $T_{\#} P = Q$. 
The concept is versus the \textit{dynamic} OT problem (the Benamou-Brenier equation) \citep{villani2009optimal,benamou2000computational}, which is less studied computationally, especially in high dimensions, with a few exceptions:

Trajectorynet \citep{tong2020trajectorynet} proposed a regularized CNF approach to learn the OT trajectory from a reference distribution $P$ -- assumed to be Gaussian (so that the KL can be estimated in the CNF) -- to a data distribution, motivated by the application of interpolating cellular distributions in single-cell data. 
Later, \citep{tong2024improving} proposed to learn the velocity field in the dynamic OT by Flow Matching, assuming that the static OT solutions on mini-batches have been pre-computed. 
In comparison, we parameterize the flow by a neural ODE and directly solve the Benamou-Brenier equation from finite samples, avoiding any pre-computation of OT couplings. 
We also allow the two endpoint distributions $P$ and $Q$ to be arbitrary, and only finite samples from each distribution are provided. 

Meanwhile, 
the rectified flow \citep{liu2022rectified},
{as a form of Flow Matching}, is closely related to the OT; however, the iterative refining approach in \citep{liu2023flow}
may not guarantee the optimality of the coupling.
\citep{kornilov2024optimal} proposed to learn dynamic OT using Flow Matching,
but the framework relied on input convex neural networks which may have limited expressiveness. In addition, each training step in \citep{kornilov2024optimal} requires costly Hessian inversion of the parametrized deep network.
Recently, \citep{shi2024diffusion,tong2024improving} proposed to use diffusion or flow models to solve the Schrödinger Bridge (SB) problem as entropy-regularized dynamic OT. 
We will experimentally compare with recent neural network OT baselines, both static and dynamic, and including SB baselines, 
in Section \ref{sec_experiments}, where our model shows better performance
e.g., on the image-to-image translation task.

\begin{figure*}[t]
\centering
\includegraphics[width=0.95\linewidth]{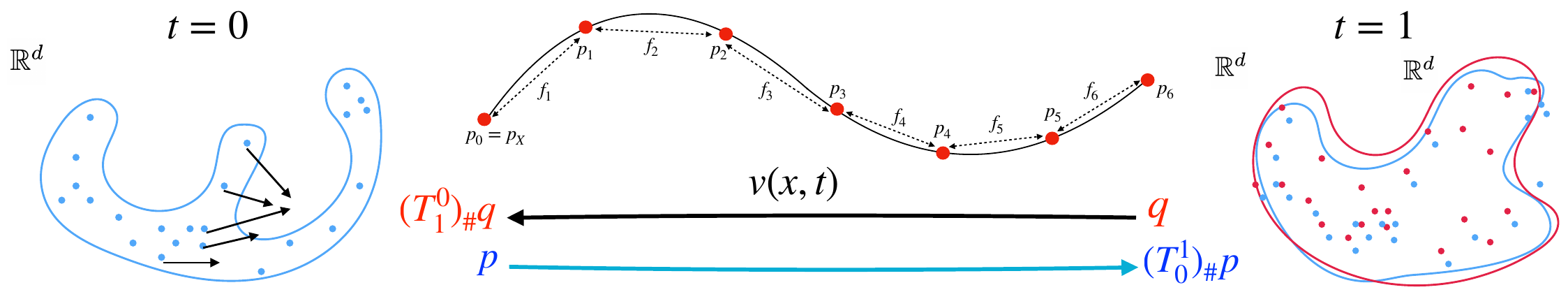}
\caption{
{The \flowname model learns the dynamic OT, 
an invertible transport map $T_0^1$ (parametrized by the velocity field $v(x,t)$)} between $P$ and $Q$ over the time interval $[0,1]$ with the least transport cost. {The push-forwarded distribution $(T_0^1)_{\#} p$ (as well as $(T_1^0)_{\#} q$, respectively) is close to the target distribution $q$ ($p$, respectively).}}
\label{fig:initialize_cartoon}
\end{figure*}

\section{Preliminaries}\label{sec_prelim}

\paragraph{Neural ODE and CNF.} 

Neural ODE \citep{chen2018neural} parameterized an ODE in $\R^d$ by a residual network. Specifically, let $x(t)$ be the solution of
\begin{equation}\label{eq:ode-Xt}
\dot{x}(t) = f (x(t), t; \theta ),  \quad x(0) \sim p.
\end{equation}
where $f(x,t; \theta)$ is a velocity field parameterized by the neural network.
Since we impose a distribution $P$ on the initial value $x(0)$, the value of $x(t)$ at any $t$ also observes a distribution $p(x,t)$ (though $x(t)$ is deterministic given $x(0)$).
In other words, $p(\cdot, t)= (T_t)_\# p$, where $T_t$ is the solution map of the ODE, namely
$T_t (x) = x + \int_0^t f(x(s),s; \theta)ds$, 
$x(0) = x$. 
%
In the context of CNF \citep{nflow_review}, the training of the flow network $f (x, t; \theta )$ is to minimize the KL divergence between the terminal density $p(x, T)$ at some $T$ and a target density $p_Z$ which is the normal distribution. The computation of the objective relies on the expression of normal density and can be estimated on finite samples of $x(0)$ drawn from $p$.

\paragraph{Dynamic OT.} 
The Benamou-Brenier equation below provides the dynamic formulation of OT  \citep{villani2009optimal,benamou2000computational}
\begin{equation}\label{eq:Benamou-Brenier}
\begin{split}
&\inf_{\rho, v}     \mathcal{T} :=  \int_0^1 \mathbb{E}_{x(t) \sim \rho(\cdot,t)} \| v(x(t),t) \|^2 dt  \\
& s.t. ~~\partial_t \rho + \nabla \cdot (\rho  v) = 0, 
   ~~  \rho(\cdot,0) = p,  ~~ \rho(\cdot,1) = q,    
\end{split}
\end{equation}
where $v(x,t)$ is a velocity field and $\rho(x,t)$ is the probability mass at time $t$ satisfying the continuity equation with $v$. 
The action $\mathcal{T}$ is the transport cost. 
Under regularity conditions of $p$, $q$, the minimum $\mathcal{T}$ in \eqref{eq:Benamou-Brenier} equals the squared Wasserstein-2 distance between $p$ and $q$, 
and the minimizer $v(x,t)$ can be interpreted as the optimal control of the transport problem.

\section{Learning Dynamic OT by Q-flow Network}\label{sec_flow}

We first introduce the formulation and training objective in Section \ref{subsec:flow-obj}.
The training technique consists of the end-to-end refinement (Section \ref{subsec:flow-refine}) 
and constructing the initial flow (Section \ref{subsec:flow-initial}). 

\subsection{Formulation and Training Objective}
\label{subsec:flow-obj}

Given two sets of  samples $\boldsymbol{X} =\{X_i\}_{i=1}^N$ and $\boldsymbol{\tilde{X}}=\{\tilde{X}_j\}_{j=1}^{M}$, where $X_i \sim P$ and $\tilde{X}_j \sim Q$ i.i.d.,
we train a neural ODE model $f(x,t;\theta)$ \eqref{eq:ode-Xt} {to represent $v(x,t)$ and solve the dynamic OT \eqref{eq:Benamou-Brenier}.}
Our formulation is symmetric from $P$ to $Q$ and vice versa, and the training objective is formally 
\begin{equation}
\min_\theta ~~ (\calL^{  P\rightarrow Q}(\theta) + \calL^{  Q\rightarrow P}(\theta)),
\label{tigheMtoL1}
\end{equation}
where we call $P\rightarrow Q$ the forward direction and $Q \rightarrow P$ the reverse direction.
The bi-directional training naturally follows the symmetry of the dynamic OT problem (Remark \ref{rk:bi-directional}).

The dynamic OT \eqref{eq:Benamou-Brenier} on time $[0,1]$ has two terminal conditions, which we propose to relax by a KL divergence loss  (see, e.g., \citep{ruthotto2020machine}).
Then the training loss in the forward direction is
\begin{equation}\label{tigheMtoL}
\calL^{  P\rightarrow Q}(\theta) = \calL_{\rm KL}^{  P\rightarrow Q}(\theta) + \gamma \calL_{T}^{  P\rightarrow Q}(\theta),
\end{equation}
where the first term $ \calL_{\rm KL}^{  P\rightarrow Q}(\theta)$ represents the relaxed terminal condition 
and the second term $\calL_{T}^{  P\rightarrow Q}(\theta)$ is the Wasserstein-2 transport cost to be detailed below;
$\gamma > 0$ is a weight parameter, and with small $\gamma$ the terminal condition is enforced. {$\calL^{  Q\rightarrow P}(\theta)$ is similarly defined, see more below.} 
{We now derive the finite-sample form of each term in \eqref{tigheMtoL}.}

\paragraph{KL loss.} 
We specify the first term $\calL_{\rm KL}^{  P\rightarrow Q} $ in loss \eqref{tigheMtoL}.
We define the solution mapping of \eqref{eq:ode-Xt} from $s$ to $t$  as
\begin{equation}\label{eq:transport_map}
    T_s^t (x; \theta)=x(s) + \int_{s}^{t} f(x(t'),t';\theta)dt',
\end{equation}
which is also parameterized by $\theta$, and we may omit the dependence below. 
By the continuity equation in \eqref{eq:Benamou-Brenier}, $\rho(\cdot, t) = (T_0^t)_\# p$. 
The terminal condition $\rho(\cdot,1)=q$ is relaxed by minimizing 
\[
{\rm KL} ( p_1 || q) = \E_{x \sim p_1} \log (p_1(x)/q(x) ), 
\quad p_1 := (T_0^1)_\# p.
\]
The expectation $\E_{x \sim p_1} $ is estimated by the sample average over $(X_1)_i$ which observes density $p_1$ i.i.d., 
where $(X_1)_i:= T_0^1(X_i)$ is computed by integrating the neural ODE from time 0 to 1.

It remains to have an estimator of $\log (p_1/q)$ to compute ${\rm KL}( p_1 || q)$. As neither $P$ nor $Q$ is assumed to have a known density, we cannot use the change-of-variable technique from normalizing flow to estimate the KL. 
We propose to train a logistic classification network 
$c_1( x; \varphi_{c}): \mathbb R^d \rightarrow \mathbb R$ with parameters $\varphi_{c}$,
which resembles the training of discriminators in GAN \citep{goodfellow2014generative}.
The inner-loop training of $c_1$ is by 
\begin{equation}\label{rnet_loss}
\min_{\varphi_c} 
 \frac 1 N \sum_{i=1}^N  \log(1+e^{ c_1 ( T_0^1(X_i;\theta); \varphi_c) } ) 
 + \frac 1 M \sum_{j=1}^{M} \log(1+ e^{-c_1(\widetilde{X}_j; \varphi_c)}).
\end{equation}
The functional optimal $c_1^*$ of the population version of loss \eqref{rnet_loss} equals $\log (q/p_1)$ by direct computation,
and as a result, ${\rm KL} ( p_1 || q) =-  \E_{x \sim p_1} c_1^*(x)  $.
Now take the trained classification network $c_1$ with parameter $\hat \varphi_c$, we can estimate the finite sample KL loss as
\begin{equation}\label{eq:min_KL}
 \calL_{\rm KL}^{  P\rightarrow Q}(\theta) = - \frac{1}{N} \sum_{i=1}^N c_1( T_0^1(X_i;\theta) ; \hat{\varphi}_c),
\end{equation}
where $ \hat{\varphi}_c$ is the computed minimizer of \eqref{rnet_loss} solved by inner loops.
In practice, we will first initialize the flow such that when minimizing  \eqref{rnet_loss}, the two densities $ p_1=(T_0^1)_\# p $ and $q$ are close,
which makes the training of the classification net more efficient.

\paragraph{Wasserstein ($W_2$) {loss}.} 
We specify the second term $\mathcal{L}_{T}^{P\rightarrow Q} (\theta)$ in the loss \eqref{tigheMtoL}.
To compute the transport cost $\mathcal{T}$ in \eqref{eq:Benamou-Brenier}, 
we use a time grid on $[0,1]$ as $0=t_0<t_1<\ldots<t_K=1$.
The choice of the time grid is algorithmic,
see more details in Section \ref{subsec:flow-refine}. 
Defining $h_k=t_k-t_{k-1}$, and $X_i(t;\theta) := T_0^{t} (X_i; \theta)$, 
we write the $W_2$ loss as
\begin{equation}\label{eq:min_W2}
\hspace{-10pt} \mathcal{L}_{T}^{P\rightarrow Q} (\theta)
    = \frac 1N \sum_{i=1}^N 
    	 \sum_{k=1}^K \frac{\|
	 X_i(t_{k};\theta) -  X_i(t_{k-1};\theta)
	 \|^2}{h_k}  ,
\end{equation}
which appears as a finite-difference approximation of $\mathcal{T}$ along time. 
Meanwhile, 
since (omitting dependence on $\theta$) 
$X_i(t_{k}) -  X_i(t_{k-1}) =  T_{t_{k-1}}^{t_k} (  X_i(t_{k-1}) )$,
the population form of \eqref{eq:min_W2}, that is,
$\sum_{k=1}^K \E_{x  \sim \rho( \cdot, t_{k-1}) } 
\| T_{t_{k-1}}^{t_k} ( x; \theta ) \|^2/ h_k$,
at optimum can be interpreted as the discrete-time summed (square) Wasserstein-2 distance, that is,
$\sum_{k=1}^K 
  { W_2( \rho( \cdot, t_{k-1} ), \rho( \cdot, t_{k} ) )^2 } / h_k
$,   
see \citep{xu2022iGNN}.
 The $W_2$ loss encourages a smooth flow from $P$ to $Q$ with a small transport cost, which also guarantees the invertibility of the model in practice 
 when the trained flow approximates the optimal flow in  \eqref{eq:Benamou-Brenier}.
  
\paragraph{Training in both directions.}
The formulation in the reverse direction is similar, where we transport $Q$-samples $\tilde{X}_j$ from time 1 to 0 using the same neural ODE integrated in reverse time. 
Specifically, 
$
\calL^{  Q\rightarrow P}(\theta) = \calL_{\rm KL}^{  Q\rightarrow P}(\theta) + \gamma \calL_{T}^{  Q\rightarrow P}(\theta)$,
and
$
  \mathcal{L}_{\rm{KL}}^{Q\rightarrow P}(\theta)
    =  - \frac{1}{M} \sum_{j=1}^{M} \tilde{c}_0( T_1^0( \tilde{X}_j; \theta); \hat{\varphi}_{\tilde{c}}  )$,
where $ \hat{\varphi}_{\tilde{c}} $ is obtained by inner-loop training of 
another classification net $\tilde{c}_0(x, \varphi_{\tilde{c}} )$ with parameters $\varphi_{\tilde{c}}$ via
\begin{equation}
\label{rnet_loss_QP}
    \min_{ \varphi_{\tilde{c}} } 
 \frac 1M \sum_{j=1}^M \log(1+e^{ \tilde{c}_0 ( T_1^0( \tilde{X}_j;\theta); \varphi_{\tilde{c}} ) } )
+ \frac 1N \sum_{i=1}^{N} \log(1+ e^{- \tilde{c}_0( {X}_i; \varphi_{\tilde{c}} )} ).
\end{equation}
The reverse-time $W_2$ loss is
\[
   \mathcal{L}_{T}^{Q \rightarrow P} (\theta)
    =  \frac{1}{M} \sum_{j=1}^M
    	\left(  \sum_{k=1}^K \frac{1}{h_k}\|
	 \tilde{X}_j(t_{k-1};\theta) -  \tilde{X}_j(t_{k};\theta)
	 \|^2\right),
     \]
  where we define $\tilde{X}_j( t; \theta) := T_1^t(\tilde{X_j}; \theta)$.

\vspace{5pt}
\begin{remark}[{Symmetry of dynamic OT and bi-directional training}]\label{rk:bi-directional}
The Benamou-Brenier formula \eqref{eq:Benamou-Brenier} is ``symmetric'' with respect to $P$ and $Q$, in the sense that either setting $P$ to be at time $0$ and $Q$ at time $1$, or the other way, one will obtain the same solution -- transport with the same velocity field $v(x,t)$ that only differs in the direction of time -- under generic conditions (e.g. both $P$ and $Q$ have densities). 
This means that training in the forward or the reversed direction, although having different objectives in appearance, is aiming at the same flow solution at optimum. 
Thus, the proposed bi-directional training naturally
captures the symmetry of the dynamic OT,
and can potentially improve the accuracy when training with finitely many samples. We empirically verify the advantage over uni-directional training in Section \ref{expr:mnist}.
\end{remark}

\subsection{End-to-end Training of \flowname}\label{subsec:flow-refine}

Assuming that the \flowname net has already been initiated (see more in Section \ref{subsec:flow-initial}), 
we minimize  $\calL^{  P\rightarrow Q} $ and $\calL^{  Q\rightarrow P} $ in an alternative fashion per iteration, and the procedure is summarized in Algorithm \ref{alg:refine}.
We call this the ``refinement'' of the flow, namely to refine the flow trajectory towards the OT one from some initialization. 
The computational complexity is analyzed in Appendix \ref{app:comp-complexity}, where the computational cost is shown to scale with the size of the time grid along the flow and the sample sizes (and number of iterations).

\begin{algorithm}[!b]
\vspace{-5pt}
\caption{\flowname refinement}\label{alg:refine}
\begin{algorithmic}[1]
\INPUT 
Pre-trained initial flow network $f(x(t),t;\theta)$; training data $\boldsymbol{X} \sim P$ and $\boldsymbol{\widetilde{X}} \sim Q$; hyperparameters: $\{ \gamma, 
\{t_k\}_{k=1}^K, 
\texttt{Tot}, 
E,
E_0, E_{\rm in}\}$.
\OUTPUT Refined flow network $f(x(t),t;\theta)$
\FOR{Iter = $1,\ldots,\texttt{Tot}$}

\STATE (If Iter = 1) Train $c_1$ by minimizing \eqref{rnet_loss} for $E_0$ epochs.
\FOR[$\triangleright$ $P\rightarrow Q$ refinement]{epoch = $1,\ldots,E$}
\STATE Update $\theta$ of $f(x(t),t;\theta)$ by minimizing 
$\calL^{  P\rightarrow Q}$.
\STATE Update $c_1$ by minimizing \eqref{rnet_loss} for $E_{\rm in}$ epochs.
\ENDFOR

\STATE (If Iter = 1) Train $\widetilde{c}_0$ by minimizing \eqref{rnet_loss_QP} for $E_0$ epochs.
\FOR[$\triangleright$ $Q\rightarrow P$ refinement]{epoch = $1,\ldots,E$}
\STATE Update $\theta$ of $f(x(t),t;\theta)$ by minimizing 
 $\calL^{  Q\rightarrow P}$.
\STATE Update $\widetilde{c}_0$ by minimizing \eqref{rnet_loss_QP} for $E_{\rm in}$ epochs.
\ENDFOR
\ENDFOR
\end{algorithmic}
\end{algorithm}

\paragraph{Time integration of flow.}
In the losses \eqref{eq:min_KL} and \eqref{eq:min_W2}, one need to compute the transported samples $X_i(t;\theta)$ and $\tilde{X}_j(t;\theta)$ on time grid points $\{ t_k \}_{k=0}^K$. 
This calls for integrating the neural ODE on $[0,1]$, which we conduct on a fine time grid $t_{k,s}$ that divides each subinterval $[t_{k-1},t_k]$ into $S$ mini-intervals ($S$ is usually 3-5 in our experiments).  
We compute the time integration of $f(x,t;\theta)$ using 
four-stage Runge-Kutta on each mini-interval. 
The fine grid is used to ensure the numerical accuracy of ODE integration and the numerical invertibility of the \flowname net, i.e., the error of using reverse-time integration as the inverse map (see inversion errors in Table \ref{inv_err}).
To further improve efficiency, 
it is possible to first train the flow $f(x,t;\theta)$ on a {coarse} time grid to warm-start the later training on a refined grid. 
One can also adopt an adaptive time grid, e.g., by enforcing equal $W_2$ movement on each subinterval $[t_{k-1}, t_k]$, so that the representative points are more evenly distributed along the flow trajectory and the learning of the flow model may be improved \citep{xu2023normalizing}. 

\paragraph{Inner-loop training of $c_1$ and $\tilde{c}_0$.}

Given a warm-started initialization of the flow, the transported distributions $(T_0^1)_\#P \approx Q$ and  $(T_1^0)_{\#}Q \approx P$. 
The two classification nets are first trained for $E_0$ epochs before the loops of training the flow model and then updated for $E_{\rm in}$ inner-loop epochs in each outer-loop iteration. 
We empirically find that {frequently updating} $c_1$ and $\widetilde{c}_0$ in lines 5 and 10 of Algorithm \ref{alg:refine} are crucial for successful end-to-end training of \flowname net.
Specifically, as we update the flow model $f(x,t;\theta)$, the push-forwarded distributions $(T_0^1)_\#P$ and $(T_1^0)_{\#}Q$ are consequently changed. Then one will need to retrain $c_1$ and $\widetilde{c}_0$ timely {(e.g., once every several steps of training $\theta$)} to ensure an accurate estimate of the log-density ratio and consequently the KL loss.
Compared with training the flow parameter $\theta$, the computational cost of the two classification nets is light 
which allows potentially a large number of inner-loop iterations {(i.e., {frequent} updates)} if needed.

\subsection{Flow Initialization}\label{subsec:flow-initial}
We propose to initialize the \flowname net by a flow model that approximately matches the transported distributions with the target distributions in both directions (and may not necessarily minimize the transport cost). 
{The proposed end-to-end training can be viewed as a refinement of the initial flow.}
The initial flow $f(x,t;\theta)$ may be specified using prior knowledge of the problem is available. 
Generally, when only two data sets $\boldsymbol{X}, \boldsymbol{\tilde{X}}$ are given, {one can adopt various existing generative flows to obtain an initial flow.}
In this work, we consider two approaches:
(i) by a concatenation of two CNF models, 
and (ii) by distribution interpolant neural networks. 
See Appendix \ref{app:flow-init} for details. 
Any other initialization scheme is compatible with the proposed end-to-end training. 

%

\section{Infinitesimal Density Ratio Estimation (DRE)}\label{sec:dre_main}

{The DRE problem, namely estimating $\log(q(x)/p(x))$ is a fundamental task in statistical inference.
\citep{rhodes2020telescoping,choi2022density} showed that an interpolated sequence of distributions from $P$ to $Q$ can be used to compute the DRE (reviewed in more detail in Appendix \ref{sec_rnet}).}
Following the same idea,  we propose to train an additional continuous-time neural network,
called {\it \rationame net}, 
by minimizing a classification loss to distinguish distributions on neighboring time stamps along the trained \flowname net trajectory. 
We will show in Section \ref{sec:dre_examples} that using the OT trajectory provided by the trained \flowname net can benefit the accuracy of DRE.

Let $p(x,t) = (T_t)_\# p$ and $T_t$ is the transport induced by the trained \flowname net.
By the relation \eqref{eq:time_score}, we propose to parametrize the time score $ {\partial_t} \log p(x,t) $ 
by a neural network  $r(x,t; \theta_r)$ with parameter $\theta_r$, which we call the {\rationame}net,
and the network $r(x,t; \theta_r)$ is to perform DRE given the OT trajectory from a pre-trained \flowname net.
The training is by logistic classification applied to transported data distributions on consecutive time grid points:
Given a deterministic time grid $0=t_0<t_1<\ldots < t_{L}=1$
(which again is an algorithmic choice; see Section \ref{sec:rnet_algo}), we expect that the integral
\begin{equation}\label{eq:def-Rk-ratio} 
\begin{split}
R_k(x; \theta_r):&= \int_{t_{k-1}}^{t_k} r( x ,t;\theta_r) dt 
\approx 
\int_{t_{k-1}}^{t_k} \partial_t \log p ( x ,t) dt\\ &= \log ( p(x, t_k)/p(x, t_{k-1})). 
\end{split}
\end{equation}
Since logistic classification can reveal the log density ratio 
as has been used in Section \ref{subsec:flow-obj}, this suggests the loss on interval $[t_{k-1}, t_k]$ as follows, for $k=1,\cdots, L$,
\begin{equation}\label{rnet_loss_continuous}
   L_k^{P \rightarrow Q} (\theta_r) 
= \frac{1}{N} \sum_{i=1}^N \log( 1 + e^{ R_k( X_i(t_{k-1}) ;\theta_r) })
	+   \frac{1}{N} \sum_{i=1}^N \log( 1+e^{- R_k( X_i(t_{k}) ;\theta_r )} ), 
\end{equation}
where $X_i(t):= T_0^t(X_i)$ and $T_0^t$ is computed by integrating the trained \flowname net.
When $k=L$, the distribution of $X_i(t_L)$ may slightly differ from that of $Q$ due to the error in matching the terminal densities in \flowname net, and then replacing the second term in \eqref{rnet_loss_continuous} with an empirical average over the $Q$-samples $\tilde{X}_j$ may be beneficial. 
In the reverse direction, define $\tilde{X}_j(t):= T_1^t( \tilde{X}_j)$,
we similarly have
$
L_k^{Q \rightarrow P} (\theta_r) 
= \frac{1}{M} \sum_{j=1}^M \log( 1 + e^{ R_k( \tilde{X}_j(t_{k-1}) ;\theta_r) }) 
	+  \frac{1}{M} \sum_{j=1}^M \log( 1+e^{- R_k( \tilde{X}_j(t_{k}) ;\theta_r )} )$,
and when $k=1$, we replace the 1st term with an empirical average over the $P$-samples $X_i$. 
The training of the  \rationame net is by 
\begin{equation}\label{rnet_loss_continuous_joint}
    \min_{\theta_r} 
    \sum_{k=1}^{L} {L}_k^{P \rightarrow Q} (\theta_r)
    + {L}_k^{Q \rightarrow P} (\theta_r).
 \end{equation}
When trained successfully, 
{the integral $\int_s^t r(x, t'; \theta_r)dt'$ provides an estimate of $\log ( p(x, t)/p(x,s))$ for any $s<t$ on $[0,1]$,
and in particular, 
the integral over $[0,1]$ yields the desired log density ratio $ \log (q/p)$ as shown in \eqref{eq:time_score}.}
See Algorithm \ref{alg:rnet} for more details.

\begin{table*}[!t]
\centering
\caption{OT benchmarks using Gaussian mixtures (left) and CelebA64 images (right). Metric values ($\mathcal{L}^{2}$-UVP, $\cos$) are shown in cells, with lower $\mathcal{L}^{2}$-UVP and higher $\cos$ being better. Results of MM, MMv1, MMv2, MM:R, and W2 are quoted from \citep{korotin2021do} for comparisons.}
\begin{minipage}{0.49\linewidth}
    \centering
    \subcaption{Gaussian mixtures}
    \label{table-hd-metrics}
    \setstretch{1.5}
    {\fontsize{30pt}{30pt}\selectfont\resizebox{\linewidth}{!}{
    \begin{tabular}{c|ccc}
    \hline
    {Dimension} & \textbf{64} & \textbf{128} & \textbf{256} \\
    \hline
    $\text{\flowname (Ours)}$ & \textbf{(4.00, 0.98)} & \textbf{(2.12, 0.99)} & \textbf{(1.97, 0.99)} \\
    OTCFM & (4.64, 0.97) & (2.78, 0.99) & (3.02, 0.98) \\ 
    MMv1 & (8.1, 0.97) & (2.2, 0.99) & (2.6, 0.99) \\
    MMv2 & (10.1, 0.96) & (3.2, 0.99) & (2.7, 0.99) \\
    W2 & (7.2, 0.97) & (2.0, 1.00) & (2.7, 1.00) \\
    \hline
    \end{tabular}}}
\end{minipage}
\begin{minipage}{0.49\linewidth}
    \centering
    \subcaption{CelebA64 images}
    \label{tab_celeba}
    \setstretch{1.5}
    {\fontsize{30pt}{30pt}\selectfont
    \resizebox{\linewidth}{!}{
    \begin{tabular}{c|ccc}
    \hline
    {Ckpt} & \textbf{Early} & \textbf{Mid} & \textbf{Late} \\
    \hline
    $\text{\flowname (Ours)}$ & \textbf{(0.87, 0.99)} & \textbf{(0.27, 0.97)} & \textbf{(0.12, 0.97)} \\
    NOT & (0.99, 0.99) & (0.34, 0.96) & (0.12, 0.96) \\
    OTCFM & (1.2, 0.99) & (0.39, 0.96) & (0.16, 0.95) \\
    MM & (2.2, 0.98) & (0.9, 0.90) & (0.53, 0.87) \\
    MM:R & (1.4, 0.99) & (0.4, 0.96) & (0.22, 0.94) \\
    \hline
    \end{tabular}}}
\end{minipage}
\end{table*}

\section{Experiments}\label{sec_experiments}

We demonstrate the effectiveness of the proposed method on several downstream tasks: 
OT baselines (Section \ref{expr:OT_benchmark}),
image-to-image translation (Section \ref{sec:img-img}),
high dimensional DRE (Section \ref{sec:dre_examples}). 
Additional details, including hyperparameter choices and sensitivity, are provided in Appendix \ref{appendix_experiments}.
Code can be found at \url{https://github.com/hamrel-cxu/FlowOT}.

\begin{figure}[!t]
    \begin{minipage}{0.495\textwidth}
        \centering
        \includegraphics[width=\textwidth]{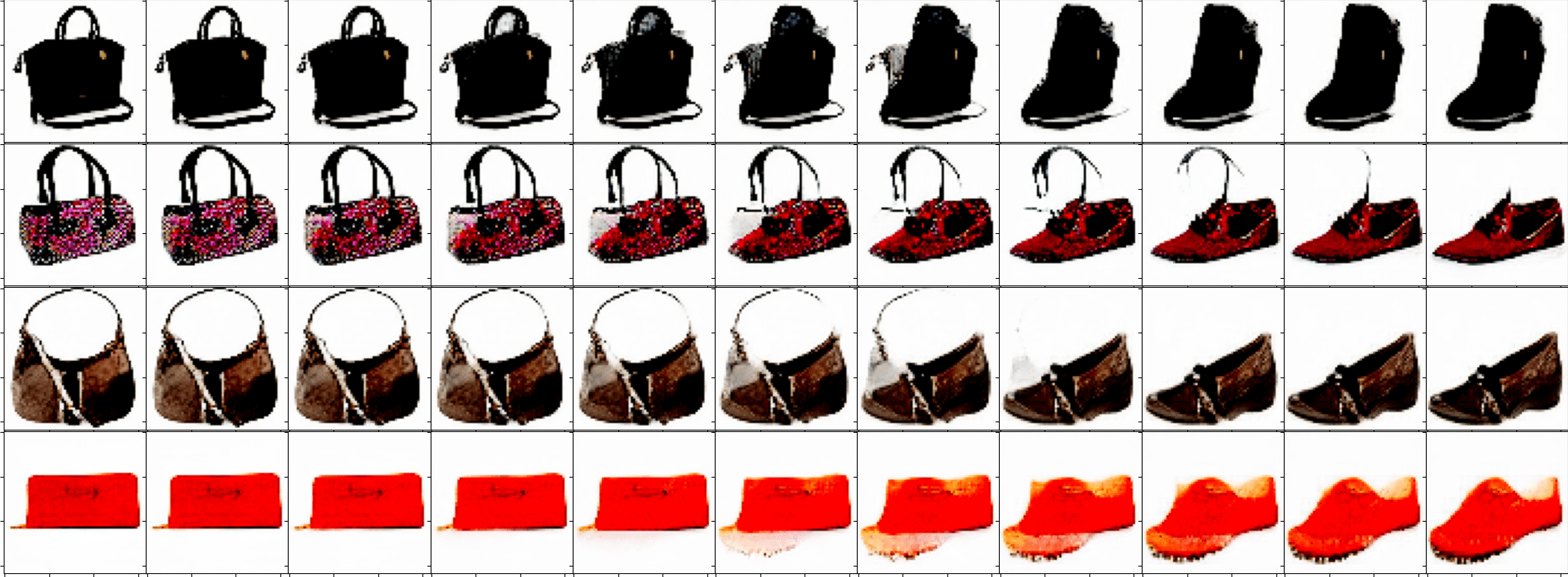}
        \subcaption{Handbag $\rightarrow$ shoes}
    \end{minipage}
    \begin{minipage}{0.495\textwidth}
        \centering
        \includegraphics[width=\textwidth]{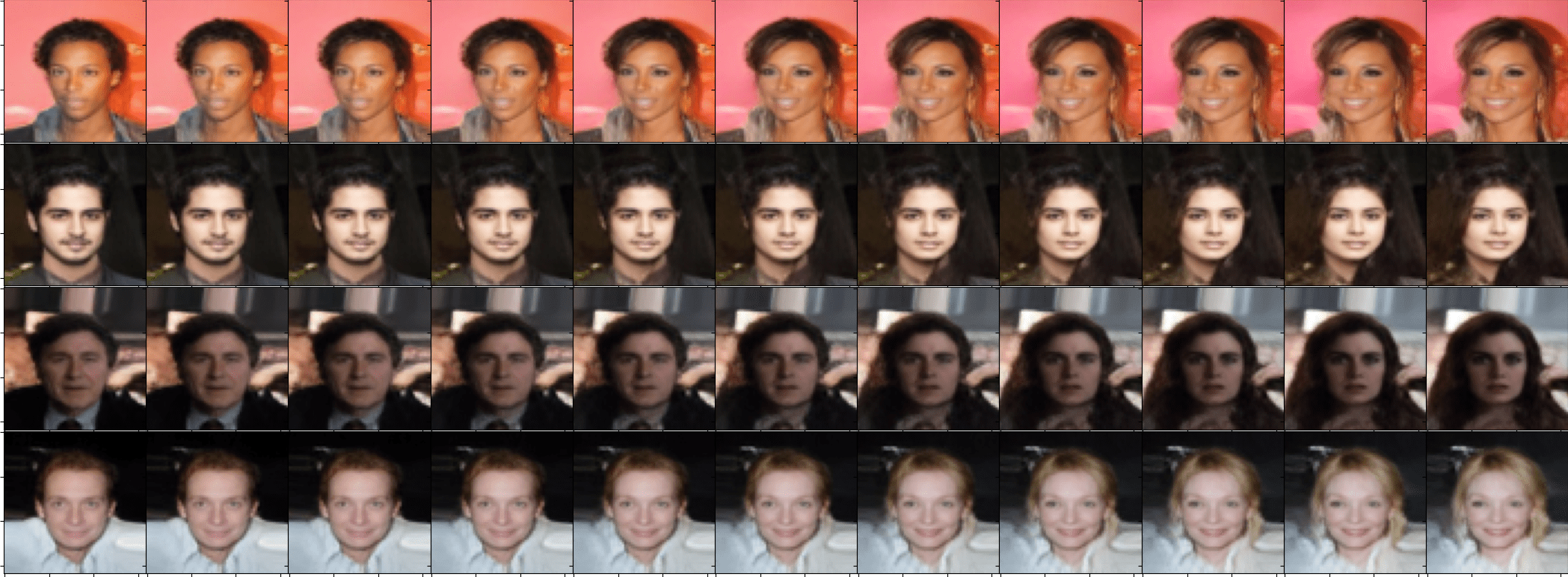}
        \subcaption{CelebA male $\rightarrow$ female}
    \end{minipage}
    \caption{Image-to-image translation: the trajectory of samples (in rows) from intermediate distributions of the Q-flow in the VAE latent space, as it pushes forward the base distribution (leftmost column) to the target distribution (rightmost column). Figure (a) shows the transition from handbag to shoe. Figure (b) shows the transition from CelebA male to female.}
    \label{fig:img_img_traj}
\end{figure}

\subsection{High-dimensional OT Baselines}\label{expr:OT_benchmark}

We compare our \flowname with popular OT baselines on the OT benchmark \citep{korotin2021do}. 
{Competing methods are $\text{MM}$ \citep{dam2019three}, $\text{MMv1}$ \citep{taghvaei20192}, $\text{MMv2}$ \citep{fan21d}, $\text{MM:R}$ \citep{makkuva2020optimal}, $\text{W2}$ \citep{korotin2021continuous},  $\text{NOT \citep{korotin2023neural}}$, and $\text{OTCFM \citep{tong2024improving}}$.}
The goal is to match outputs from a trained OT map as closely as possible with those from the ground truth in the benchmark. Two metrics, $\mathcal{L}^{2}$-UVP in \eqref{eq:l2_uvp} and $\cos$ in \eqref{eq:cos}, are used to evaluate performance, where lower $\mathcal{L}^{2}$-UVP and higher $\cos$ values indicate better performance. Details of the setup are provided in Appendix \ref{appendix_w2_benchmark}. Table \ref{table_compute_time} reports the training time of each method, where \flowname is computationally competitive.

\paragraph{High-dimensional Gaussian mixtures.} The goal is to transport between high-dimensional Gaussian mixtures optimally. We vary dimension $d \in \{64, 128, 256\}$, and in each dimension, the distribution $P$ is a mixture with three components, and $Q$ is a weighted average of two 10-component Gaussian mixtures. Table \ref{table-hd-metrics} shows that our \flowname consistently reaches lower $\mathcal{L}^{2}$-UVP and higher or equal $\cos$ than OT baselines, indicating comparable or better performance on this example.

\paragraph{CelebA64 images.} The goal is to align CelebA64 faces \citep{liu2015faceattributes} (denoted as $Q$) with faces generated by a pre-trained WGAN-QC \citep{liu2019wasserstein} generator (denoted as $P_{\text{Ckpt}}$ for different checkpoints). In particular, three checkpoints (Ckpt) of the WGAN-QC are considered (i.e., $\text{Ckpt} \in \{\text{Early, Mid, Late}\}$), where $P_{\text{Early}}$ contains generated faces that are the most blurry and faces from $P_{\text{Late}}$ are closest to true faces among the three. Quantitatively, results in Table \ref{tab_celeba} show that \flowname outperforms other OT baselines. Figure \ref{fig_celeba} in the appendix further shows high-quality faces as a result of using the trained \flowname net on samples from $P_{\text{Ckpt}}$ for different Ckpt.


\begin{table*}[!t]
    \centering
    \caption{Image-to-image translation: FID on the test sets, lower is better. FIDs of Disco GAN, Cycle GAN, and NOT are quoted from \citep{korotin2023neural}.}\label{tab:fid}
    \setstretch{1.35}
    {\fontsize{30pt}{30pt}\selectfont\resizebox{\linewidth}{!}{
    \begin{tabular}{c|cccccccccc}
    \toprule
     & \flowname (ours) & OTCFM & Re-flow & {SBCFM} & {DSBM} & {W2GN} & MM:R & Disco GAN & Cycle GAN & NOT \\
    \hline
    Handbag $\rightarrow$ shoes & \textbf{12.34} & 15.96 & 25.92 & {17.22} & {25.71} & {34.23} & 33.04 & 22.42 & 16.00 & 13.77\\
    CelebA male $\rightarrow$ female & \textbf{9.66} & 9.76 & 20.24 & {11.32} & {25.82} & {15.23} & 12.34 & 35.64 & 17.74 & 13.23\\
\bottomrule
    \end{tabular}
    }}
\end{table*}

\begin{figure*}[!t]
\vspace{10pt}
    \centering
    \begin{minipage}{\linewidth}
    \includegraphics[width=1\linewidth]{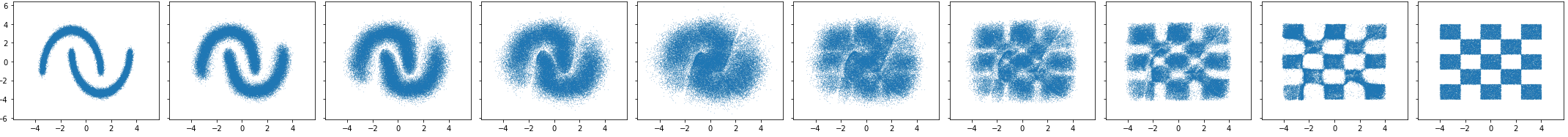}
    \subcaption{Trajectory from $P$ (two-moon) to $Q$ (checkerboard)}
    \end{minipage}
    \begin{minipage}{0.95\linewidth}
    \includegraphics[width=\linewidth]{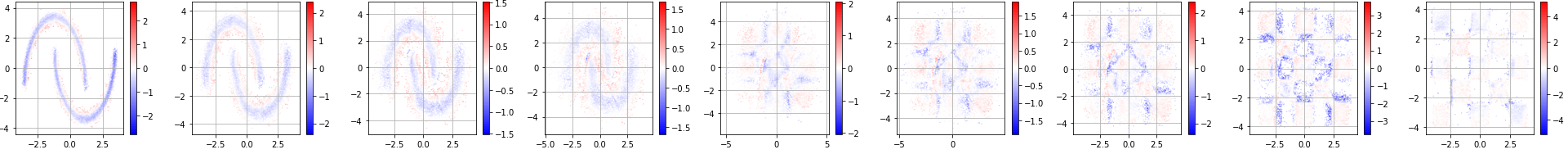}
    \subcaption{
    Estimated $\log$-ratio between $P_{t_{k-1}}$ and $P_{t_k}$ by the trained \rationame net.
    }
    \end{minipage}
    \caption{\flowname trajectory 
    and corresponding $\log$-ratio estimation. 
    \textbf{Top}: intermediate distributions by \flowname net. \textbf{Bottom}: 
    corresponding $\log$-ratio estimated by \rationame net.
    Blue color indicates negative estimates of the difference $\log (p(x,t_k)/p(x,t_{k-1}))$ evaluated at the common support of the neighboring densities.
    }
    \label{moon_to_checkerboard}
\end{figure*}

\subsection{Image-to-image Translation} \label{sec:img-img}
We use \flowname to learn the continuous-time OT between distributions of two sets of RGB images and compare with Disco GAN \citep{kim2017learning}, Cycle GAN \citep{zhu2017unpaired}, {W2GN \citep{korotin2021wasserstein}}, MM:R, NOT, Re-flow \citep{liu2023flow}, OTCFM, {SBCFM \citep{tong2024improving}, and DSBM \citep{shi2024diffusion}}. 
The first set contains handbags \citep{zhu2016generative} and shoes \citep{yu2014fine}, and the second set contains CelebA male and female images. We denote handbags/males as $P$ and shoes/females as $Q$.
We follow the setup in \citep{korotin2023neural}, where the goal of the image-to-image translation task is to conditionally generate shoe/female images by mapping images of handbag/male through our trained \flowname model. We train \flowname in the latent space of a pre-trained variational auto-encoder (VAE) on $P$ and $Q$. {For a fair comparison, we train OTCFM, Re-flow, SBCFM, DSBM, W2GN, and MM:R in the same latent space for the same number of mini-batches as \flowname net}, and all methods use models of the same size. Additional details are in Appendix \ref{app:handbag_shoes}. 

Figure \ref{fig:img_img_traj} visualizes continuous trajectories from handbags/males to shoes/females generated by the \flowname model. We find that \flowname can capture the style and color nuances of corresponding handbags/males in the generated shoes/females as the flow model continuously transforms handbag/male images. Figure \ref{fig:handbag_shoes_more} and \ref{fig:celeba_more} in the appendix show additional generated shoes/females from handbags/males, respectively.
Quantitatively, FIDs between generated and true images in the test set, as shown in Table \ref{tab:fid} indicate \flowname performs better than all baselines.
Meanwhile, as our \flowname model learns a \textit{continuous} transport map from source to target domains, it directly provides the gradual interpolation between source and target samples along the dynamic OT trajectory (Figure \ref{fig:img_img_traj}).

\vspace{-4pt}
\subsection{High-dimensional DRE
}\label{sec:dre_examples}
\vspace{-4pt}

We show the benefits of our \rationame net proposed in Section \ref{sec:dre_main} on various DRE tasks, where \rationame net leverages the flow trajectory of a trained \flowname net.
In the experiments below, we denote our method as ``\ours{}'',
and compare against three baselines of DRE in high dimensions. The baseline methods are: 
1 ratio (by training a single classification network using samples from $P$ and $Q$), 
TRE \citep{rhodes2020telescoping}, 
and \DREinf{} \citep{choi2022density}. 
We denote $P_{t_k}$ with density $p(\cdot,t_k)$ as the pushforward distribution of $P$ by the \flowname transport  over the interval $[0,t_k]$.
The set of distributions $\{P_{t_k}\}$ for $k=1,\ldots,L$ builds a bridge between $P$ and $Q$.
\begin{table*}[!t]
    \centering
    \caption{
    BPD on the energy-based modeling of MNIST (lower is better). Results for \DREinf{} are from \citep{choi2022density}, and results for one-ratio and TRE are from \citep{rhodes2020telescoping}.}
    \label{tab:bpd}
    \renewcommand{\arraystretch}{1}
   \setstretch{1.1}
    {\fontsize{40pt}{40pt}\selectfont
    \resizebox{0.85\linewidth}{!}{
    \begin{tabular}{c|cccc|cccc|cccc}
    \toprule
\textbf{Choice of $Q$} & \multicolumn{4}{c|}{RQ-NSF} & \multicolumn{4}{c|}{Copula} & \multicolumn{4}{c}{Gaussian}   \\
\hline
\textbf{Method} & \ours{} & \DREinf{} & TRE & 1 ratio & \ours{} & \DREinf{} & TRE & 1 ratio & \ours{} & \DREinf{} & TRE & 1 ratio \\
\textbf{BPD ($\downarrow$)} & \textbf{1.05} & 1.09  & 1.09 & 1.09 & \textbf{1.14} & 1.21 & 1.24 & 1.33 & \textbf{1.31} & 1.33  & 1.39 & 1.96 \\
\bottomrule
    \end{tabular}
    }}
\end{table*}
\subsubsection{Toy Data in Two-dimension}

\paragraph{Gaussian mixtures.}
We simulate $P$ and $Q$ as two Gaussian mixture models with three and two components, respectively,
see additional details in Appendix \ref{appendix_2dGMM}. 
We compare ratio estimates $\hat{r}(x)$ with the true values $r(x)$.
The results are shown in Figure \ref{fig:2dGMM}. 
We see from the top panel that the mean absolute error (MAE) of \ours{} is the smallest, and \ours{} also incurs a smaller $\max_x |\hat{r}(x)-r(x)|$. 
This is consistent with the closest resemblance of \ours{}
to the ground truth (first column) in the bottom panel.
In comparison, \DREinf{} tends to over-estimate $r(x)$ on the support of $Q$, while TRE and one ratio under-estimate $r(x)$ on the support of $P$.
As both the \DREinf{} and TRE models use the linear interpolant scheme \eqref{eq:linear_interpolate}, the result suggests the benefit of using \flowname trajectory for DRE.

\paragraph{Two-moon to and from checkerboard.}
We design two densities in $\R^2$ where $P$ represents the shape of two moons, and $Q$ represents a checkerboard; see additional details in Appendix \ref{appendix_moon_to_checkerboard}. 
For this more challenging case, 
the linear interpolation scheme \eqref{eq:linear_interpolate} creates a bridge between $P$ and $Q$ as shown in Figure \ref{fig:linear_moon_to_checkerboard_interpolate}.
The flow visually differs from the one obtained by the trained \flowname net, as shown in Figure \ref{moon_to_checkerboard}(a), and the latter is trained to minimize the transport cost.
The result of \rationame net is shown in Figure \ref{moon_to_checkerboard}(b).
The corresponding density ratio estimates of $\log p(x,t_k)-\log p(x,t_{k-1})$ visually reflect the actual differences in the two neighboring densities. 
Figure \ref{moon_to_checkerboard_PQ} additionally shows the estimates $\log q(x)-\log p(x)$ using \rationame net.

\subsubsection{High-dimensional Mutual Information 
}\label{expr:MI}

We estimate the mutual information (MI) between two correlated random variables in dimensions $d\in \{40, 80, 160, 320\}$, following the setup in \citep{rhodes2020telescoping}. 
Additional details can be found in Appendix \ref{appendix_MI}.
{
As shown in Figure \ref{fig:MI_gaussian}, the estimated MI by our method aligns well with the ground truth MI values, reaching nearly identical performance as \DREinf{} and outperforming the other two baselines.
}

\subsubsection{Energy-based Modeling of MNIST}\label{expr:mnist}

We apply our approach in evaluating and improving an energy-base model (EBM) on the MNIST dataset \citep{LeCun2005TheMD}.
We follow the prior setup in \citep{rhodes2020telescoping,choi2022density}, where $P$ is the empirical distribution of MNIST images, 
and $Q$ is the generated image distributions by three pre-trained energy-based generative models: Gaussian, Copula, and RQ-NSF \citep{durkan2019neural}. 
The performance of DRE is measured using the ``bits per dimension'' (BPD) metric in \eqref{eq:bpd}. Additional details are in Appendix \ref{appendix_MNIST}.
The results show that \ours{} reaches the improved performance in Table \ref{tab:bpd} against baselines: it consistently reaches the smallest BPD across all choices of $Q$. Meanwhile, we also note computational benefits in training: on one A100 GPU, \ours{} took approximately 8 hours to converge while \DREinf{} took approximately 33 hours. In addition, we show the trajectory of improved samples from $Q$ to $\tilde{Q}$ for RQ-NSF using the trained \flowname in Figure \ref{fig:RQ_NSF_improved_traj}. Figure \ref{fig:MNIST_improved} shows additional improved digits for all three specifications of $Q$. {Lastly, Table \ref{tab:bpd_bidir} shows the empirical benefit of bi-directional over uni-directional training of \flowname for this task.}

\section{Discussion}\label{sec:discussion}
The  \flowname model developed in this work is optimized to find the dynamic OT  transport between two distributions to learn from data samples. 
One limitation is the computational cost associated with the neural ODE training. 
To save computation, one can explore more advanced time discretization schemes, such as adaptive time grids, as well as customized neural ODE solvers \citep{shaul2024bespoke}. 
One can try to develop a simulation-free approach under the proposed framework; more efficient computation can also be achieved via progressive distillation \citep{salimans2022progressive} from the trained OT trajectory.
Meanwhile, there remain open theoretical questions, such as the theoretical guarantee of learning the OT trajectory.
{In particular, the boundary condition in the Benamou-Brenier equation is handled by KL divergences in the current model, and it would be helpful to analyze the consistency of such an approach, especially under a finite-sample scenario. 
It would also be interesting to explore alternative distribution divergences (other than KL) to handle the two endpoint distributions and analyze them under a more general framework.}
For the empirical results, extending to a broader class of applications and additional real datasets will be useful. 

\vspace{-7pt}
\section*{Acknowledgments}
\vspace{-7pt}
The work is supported by NSF DMS-2134037. 
C.X. and Y.X. are partially supported by an
NSF CAREER CCF-1650913, CMMI-2015787, CMMI-2112533, DMS-1938106, DMS-1830210,
and the Coca-Cola Foundation. 
X.C. is also partially supported by NSF DMS-2237842 and Simons Foundation MPS-MODL-00814643.

\bibliography{arxiv_references}
\bibliographystyle{plainnat}

\appendix


\setcounter{table}{0}
\setcounter{figure}{0}
\setcounter{equation}{0}

\renewcommand{\thetable}{A.\arabic{table}}
\renewcommand{\thefigure}{A.\arabic{figure}}
\renewcommand{\theequation}{A.\arabic{equation}}
\renewcommand{\thealgorithm}{A.\arabic{algorithm}}

\section{Infinitesimal density ratio estimation}\label{sec_rnet}

We first introduce related works of DRE (Section \ref{appendix:dre_related}) and DRE preliminaries (Section \ref{appendix:dre_prelim}). We then present the complete algorithm for our proposed \rationame net in Section \ref{sec:rnet_algo}.

\subsection{DRE literature} \label{appendix:dre_related}
Density ratio estimation between distributions $P$ and $Q$ is a fundamental problem in statistics and machine learning \citep{meng1996simulating,sugiyama2012density,choi2021featurized}. It has direct applications in important fields such as importance sampling \citep{neal2001annealed}, change-point detection \citep{kawahara2009change}, outlier detection \citep{kato2021non}, mutual information estimation \citep{belghazi2018mutual}, etc. Various techniques have been developed, including probabilistic classification \citep{qin1998inferences,bickel2009discriminative}, moment matching \citep{gretton2009covariate}, density matching \citep{sugiyama2008direct}, etc. 
Deep NN models have been leveraged in classification approach \citep{moustakides2019training} due to their expressive power. 
However, as has been pointed out in \citep{rhodes2020telescoping}, the estimation accuracy by a single classification may degrade when $P$ and $Q$ differ significantly. 

To overcome this issue, \citep{rhodes2020telescoping} introduced a telescopic DRE approach by constructing intermediate distributions to bridge between $P$ and $Q$. 
\citep{choi2022density} further proposed to train an infinitesimal, continuous-time ratio net via the so-called time score matching. 
Despite their improvement over the prior classification methods, both approaches rely on an unoptimal construction of the intermediate distributions between $P$ and $Q$.
In contrast, our proposed \flowname network leverages the expressiveness of deep networks to construct the intermediate distributions by the continuous-time flow transport, and the flow trajectory is regularized to minimize the transport cost in dynamic OT. 
The model empirically improves the DRE accuracy (see Section \ref{sec_experiments}).
In computation, \citep{choi2022density} applies score matching to compute the infinitesimal change of log-density.
The proposed \rationame net is based on classification loss training using a fixed time grid which avoids score matching and is computationally lighter (Section \ref{sec:rnet_algo}).

\subsection{Telescopic and infinitesimal DRE preliminaries} \label{appendix:dre_prelim}

To circumvent the problem of DRE distinctly different $p$ and $q$, the {\it telescopic DRE} \citep{rhodes2020telescoping} proposes to ``bridge'' the two densities by a sequence of intermediate densities $p_k$, $k=0, \cdots, L$, where $p_0 = p$ and $p_L = q$. The consecutive pairs of $(p_{k}, p_{k+1})$ are chosen to be close so that the DRE can be computed more accurately, and then by 
\begin{equation}\label{eq:telescopic-dre}
\log ({q}(x)/{p}(x)) = \log p_L(x) - \log p_0(x) = \sum_{k=0}^{L-1} \log p_{{k+1}}(x) -\log p_{k}(x),
\end{equation}
the log-density ratio between $q$ and $p$ can be computed with improved accuracy than a one-step DRE.
The {\it infinitesmal DRE} \citep{choi2022density} 
considers a time continuity version of \eqref{eq:telescopic-dre}. 
 Specifically, suppose the time-parameterized density $p(x,t)$ is differentiable on  $t \in [0,1]$ with $p(x,0)=p$ and $p(x,1) =q$, then
 \begin{equation}\label{eq:time_score}
 \log (q(x)/p(x)) = \log p(x, 1) -\log p( x, 0)  = \int_0^1 \partial_t \log p(x, t) dt.
\end{equation}
The quantity $ {\partial_t} \log p(x,t) $ was called the ``time score'' and can be parameterized by a neural network.

\begin{figure}
    \begin{minipage}{\linewidth}
        \begin{algorithm}[H]
        \caption{Infinitesimal DRE training via \rationame net}\label{alg:rnet}
        \begin{algorithmic}[1]
        \INPUT Training samples $\boldsymbol{X} \sim P$ and $\boldsymbol{\widetilde{X}} \sim Q$; 
        pre-trained \flowname net $f(x(t),t;\theta)$; 
        hyperparameters: \{$\{t_k\}_{k=1}^{L}$, \texttt{Tot\_iter}\}
        \OUTPUT Trained network $r(x,t;\theta_r)$.
        \FOR{$k=1,\ldots,L-1$}
        \STATE Obtain $\{X_i(t_k)\}_{i=1}^N, \{\tilde{X}_j(t_k)\}_{j=1}^M$ by transporting all training samples $\{\boldsymbol{X},\boldsymbol{\tilde{X}}\}$ using the given \flowname net $f(x,t;\theta)$
        \ENDFOR
        \FOR{Iter = $1,\ldots,\texttt{Tot\_iter}$}
        \STATE Draw mini-batches of samples from $\{X_i(t_k)\}_{i=1}^N, \{\tilde{X}_j(t_k)\}_{j=1}^M$.
        \STATE Train $\theta_r$ upon minimizing \eqref{rnet_loss_continuous_joint}.
        \ENDFOR
        \end{algorithmic}
        \end{algorithm}
    \end{minipage}
\end{figure}

We use a trained \flowname network $f(x,t;\theta)$ for infinitesimal DRE as a focused application.

\subsection{Algorithm and computational complexity}\label{sec:rnet_algo}

Algorithm \ref{alg:rnet} presents the complete \rationame algorithm. We use an evenly spaced time grid $t_k=k/L$ in all experiments. 
In practice, one can also progressively refine the time grid in training, starting from a coarse grid to train a \rationame net $r(x,t;\theta_r) $ and use it as a warm start for training the network parameter $\theta_r$ on a refined grid. 
When the time grid is fixed, it allows us to compute the transported samples $\{X_i(t_k)\}_{i=1}^N, \{\tilde{X}_j(t_k)\}_{j=1}^M$ on all $t_k$ once before the training loops of \rationame net (line 1-3).
This part takes ${O}(8 KS (M+N))$ function evaluations of the pre-trained \flowname net $f(x,t;\theta)$. 
Suppose the training loops of lines 4-6 conduct $E$ epochs in total. 
Assume each time integral in $R_k$ \eqref{eq:def-Rk-ratio} is computed by a fixed-grid four-stage Runge-Kutta method, 
then ${O}(4 L E(M+N))$ function evaluations of $r(x,t; \theta_r)$ is needed to compute the overall loss \eqref{rnet_loss_continuous_joint}.

\section{Additional experimental details}\label{appendix_experiments}

When training all networks, we use the Adam optimizer \citep{adam}. Unless otherwise specified, we use an initial learning rate of 0.001.

To provide an initialized flow, we either train a continuous-time flow via \citep{albergo2023building} in Sections \ref{expr:OT_benchmark}-\ref{sec:img-img} or concatenate two CNFs (each trained via \citep{xu2023normalizing}) in Section \ref{sec:dre_examples}.

\subsection{High-dimensional OT baselines}\label{appendix_w2_benchmark}
We summarize the setup and comparison metrics based on \citep{korotin2021do}. 
To assess the quality of a trained transport map $\hat{T}: \R^d \rightarrow \R^d$ from $P$ to $Q$ against a ground truth $T^*: \R^d\rightarrow \R^d$, the authors use two metrics: \textit{unexplained variance percentage} ($\mathcal{L}^2$-UVP) \citep{korotin2021wasserstein} and \textit{cosine similarity} ($\cos$), which are defined as
\begin{align}
   \mathcal{L}^2\text{-UVP} &= 100\cdot \frac{\mathbb{E}_{x\sim P} \|\hat{T}(x)-T^*(x)\|_2^2}{\text{Var}(Q)}\%. \label{eq:l2_uvp}\\
   \cos & = \frac{\mathbb{E}_{x\sim P}\left\langle\hat{T}(x)-x, T^*(x)-x\right\rangle_{\ell_2}}{\mathbb{E}_{x\sim P}\|\hat{T}(x)-x\|_2 \cdot\mathbb{E}_{x\sim P}\|T^*(x)-x\|_2}. \label{eq:cos}
\end{align}
The metrics are evaluated using $2^{14}$ random samples from $P$.

\begin{figure}[!t]
     \centering
     \begin{minipage}{\textwidth}
         \includegraphics[width=\linewidth]{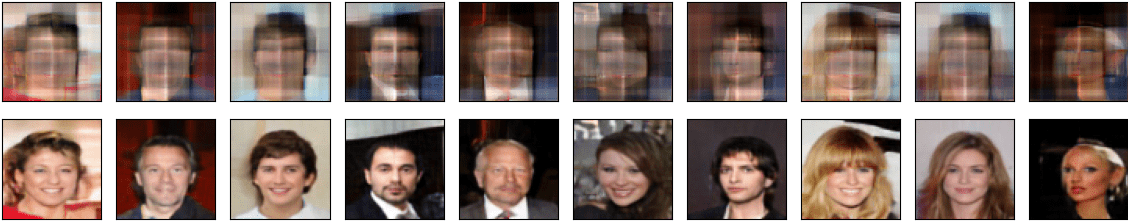}
         \subcaption{Fitted \flowname on $P_{\text{Early}} ~(\text{top}) \rightarrow Q ~(\text{bottom})$.}
     \end{minipage}
     \begin{minipage}{\textwidth}
        \includegraphics[width=\linewidth]{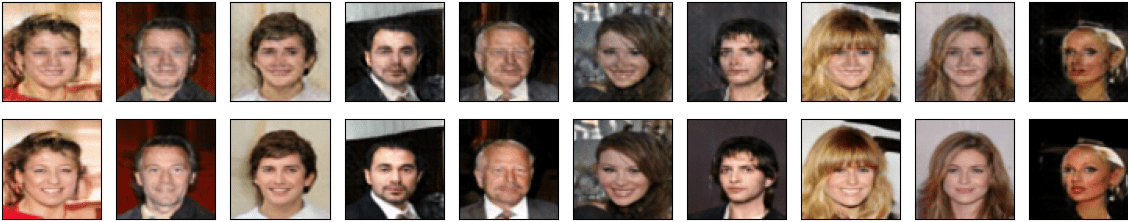}
         \subcaption{Fitted \flowname on $P_{\text{Mid}} ~(\text{top}) \rightarrow Q ~(\text{bottom})$.}
     \end{minipage}
     \begin{minipage}{\textwidth}
         \includegraphics[width=\linewidth]{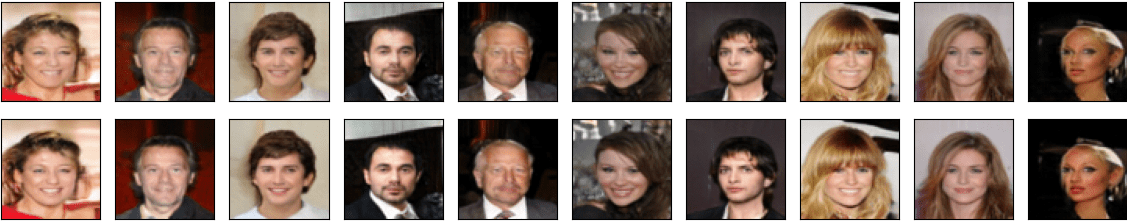}
         \subcaption{Fitted \flowname on $P_{\text{Late}} ~(\text{top}) \rightarrow Q ~(\text{bottom})$.}
     \end{minipage}
    \centering
    \caption{Fitted \flowname between $P_{\text{Ckpt}}$ for $\text{Ckpt} \in \{\text{Early, Mid, Late}\}$ and $Q$.
    In each sub-figure, the first row contains random $x\sim P_{\text{Ckpt}}$ and the second row contains generated faces $T(x)$ using the trained Q-flow net $T$.}
    \label{fig_celeba}
\end{figure}

\paragraph{High-dimensional Gaussian mixtures.} In each dimension $d$, the distribution $P$ is a mixture of three Gaussians. The distribution $Q$ is constructed as follows: first, the authors construct two Gaussian mixtures $Q_1$ and $Q_2$ with 10 components. Then, they train approximate transport maps $\nabla \psi_i \# P\approx Q_i$, where $\psi_i$ is trained via \citep{korotin2021continuous}. The target $Q$ is obtained as $Q=\frac{1}{2}(\nabla \psi_1+\nabla \psi_2)\# P$.

In Q-flow, the flow architecture $f(x(t),t;\theta)$ consists of fully connected layers of dimensions $2d-4d-8d-4d-2d$ with ReLU activation. The initialization of the flow is done by the method of Interflow \citep{albergo2023building}, where at each step, we draw random batches of 2048 samples from $P$ and $Q$. We trained the initialized flow for 50K steps. To apply Algorithm \ref{alg:refine}, we let $\gamma=0.1$, $t_k=k/5$ for $k=0,\ldots,5$, and \texttt{Tot}=1. The classifiers $c_1$ and $\tilde{c}_0$ consist of fully-connected layers of dimensions $4d-4d-4d-4d$ with ReLU activation.
These classifiers are initially trained for 10000 batches with batch size 2048. We train flow parameters $\theta$ for 10000 batches in every iteration with a batch size of 2048, and we update the training of $c_1$ and $\tilde{c}_0$ every 10 batches of training $\theta$ to train them for 10 inner-loop batches.

\paragraph{CelebA64 images \citep{liu2015faceattributes}} The authors first fit 2 generative models using WGAN-QC \citep{liu2019wasserstein} on the CelebA dataset. They then pick intermediate training checkpoints to product continuous measures $P^k_{\text{Early}},P^k_{\text{Mid}},P^k_{\text{Late}}$ for these 2 models ($k=1,2$). Then, for each $k\in \{1,2\}$ and $\text{Ckpt} \in \{\text{Early, Mid, Late}\}$, they use \citep{korotin2021continuous} to fit an approximate transport map $\nabla\psi^k_{\text{Ckpt}}$ such that $\nabla\psi^k_{\text{Ckpt}} \# Q \approx P^k_{\text{Ckpt}}$. The distributions $P_{\text{Ckpt}}$ are then obtained as $P_{\text{Ckpt}}=\frac{1}{2}(\nabla \psi^1_{\text{Ckpt}} + \nabla \psi^2_{\text{Ckpt}}) \# Q.$

In Q-flow, the flow architecture $f(x(t),t;\theta)$ consists of convolutional layers of dimensions 64-128-128-256-256-512-512, followed by convolutional transpose layers whose filters mirror the convolutional layers. The kernel sizes are 3-4-3-4-3-4-3-3-4-3-4-3-4-3 with strides 1-2-1-2-1-2-1-1-2-1-2-1-2-1. We use the ReLU activation. The initialization of the flow is done by the method of Interflow \citep{albergo2023building} for 30K steps with 128 batch size. To apply Algorithm \ref{alg:refine}, we let $\gamma=1$, $t_k=k/3$ for $k=0,\ldots,3$, and \texttt{Tot}=1. The architecture of the classifier networks $c_1$ and $\tilde{c}_0$ are ResNets used in WGAN-QC \citep{liu2019wasserstein}.
These classifiers are initially trained for 5000 batches with batch size 128. We train flow parameters $\theta$ for 10000 batches in every iteration with a batch size of 128, and we update the training of $c_1$ and $\tilde{c}_0$ every 5 batches of training $\theta$ to train them for 1 inner-loop batch. 

\paragraph{Training time.}
Table \ref{table_compute_time} reports the training time of different methods on these examples; for a given method, the time is consistent across all examples in a given table because of the same training procedure and hyper-parameters on these examples. 
The training time of \flowname does not include the time to pre-train an initialized flow, which is an input to Algorithm \ref{alg:refine}. Nevertheless, pre-training is light---on examples in Table \ref{table-hd-metrics}, pre-training takes roughly 7 minutes (9\% of \flowname training time) and on examples in Table \ref{fig_celeba}, pre-training takes roughly 18 minutes (12\% of \flowname training time). Thus considering the pre-training will not change the comparison much under this setting.

\begin{table}[!t]
    \centering
    \caption{Wall-clock training time to reach the performance in Table \ref{table-hd-metrics} and Table \ref{tab_celeba}. The unit is in hours on a single A100 GPU.}
    \label{table_compute_time}
    \renewcommand{\arraystretch}{1}
    \setstretch{1.25}
    {\fontsize{30pt}{30pt}\selectfont
    \resizebox{0.35\linewidth}{!}{
    \begin{tabular}{ccccc}
    \toprule
    \multicolumn{5}{c}{\textbf{Table \ref{table-hd-metrics} time}} \\
    \hline
    \flowname & OTCFM & MMv1 & MMv2 & W2\\
    1.25 & 1.25  & 2.5 & 1.25 & 1.25 \\
    \bottomrule
    \end{tabular}
    }}
    {\fontsize{30pt}{30pt}\selectfont
    \resizebox{0.35\linewidth}{!}{
    \begin{tabular}{ccccc}
    \toprule
    \multicolumn{5}{c}{\textbf{Table \ref{tab_celeba} time}} \\
    \hline
    \flowname & NOT & OTCFM & MM & MM:R  \\
    2.5 & 2.75 & 2.5 & 2 & 1.75 \\
    \bottomrule
    \end{tabular}
    }}
\end{table}

\begin{figure}[!b]
    \begin{minipage}{\textwidth}
        \includegraphics[width=\linewidth]{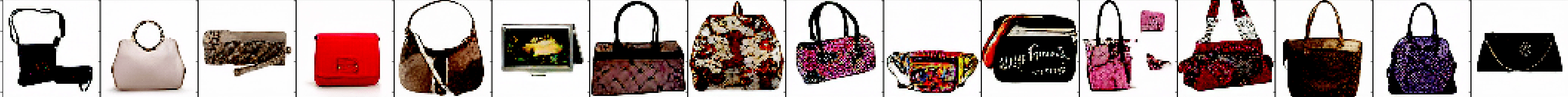}
        \subcaption{Test handbag images from $P$}
    \end{minipage}
    \begin{minipage}{\textwidth}
        \includegraphics[width=\linewidth]{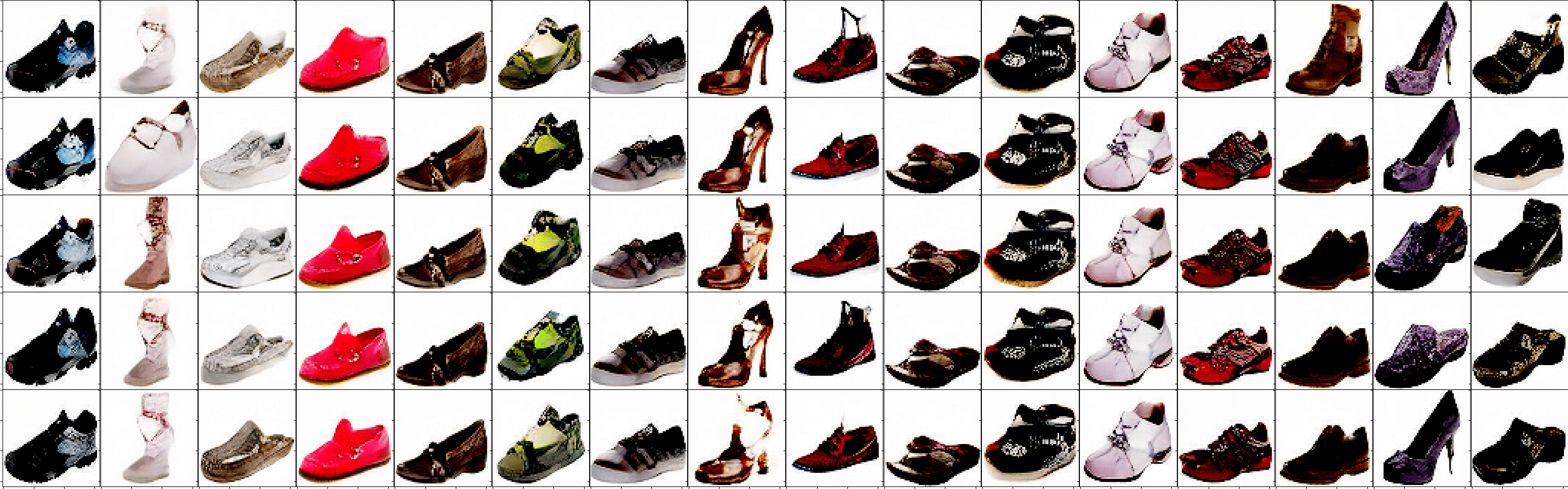}
        \subcaption{Different generated shoe images by \flowname model}
    \end{minipage}
    \caption{Additional test images of handbag (in true $P$, figure (a)) and corresponding generated shoe images by \flowname model (in generated $Q$, figure (b)). To generate different shoes for a given handbag, we sample random latent codes given by the VAE, map them through the trained \flowname model, and decode them back through the VAE decoder to visualize different generated shoes in the pixel space.}
    \label{fig:handbag_shoes_more}
\end{figure}

\begin{figure}[!t]
    \begin{minipage}{\textwidth}
        \includegraphics[width=\linewidth]{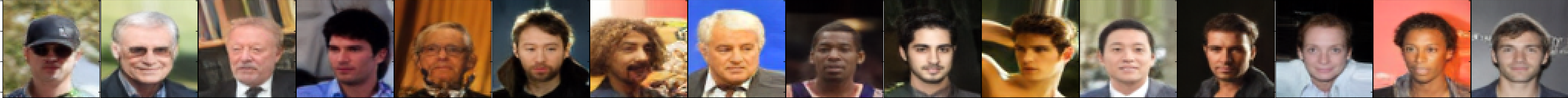}
        \subcaption{Test male images from $P$}
    \end{minipage}
    \begin{minipage}{\textwidth}
        \includegraphics[width=\linewidth]{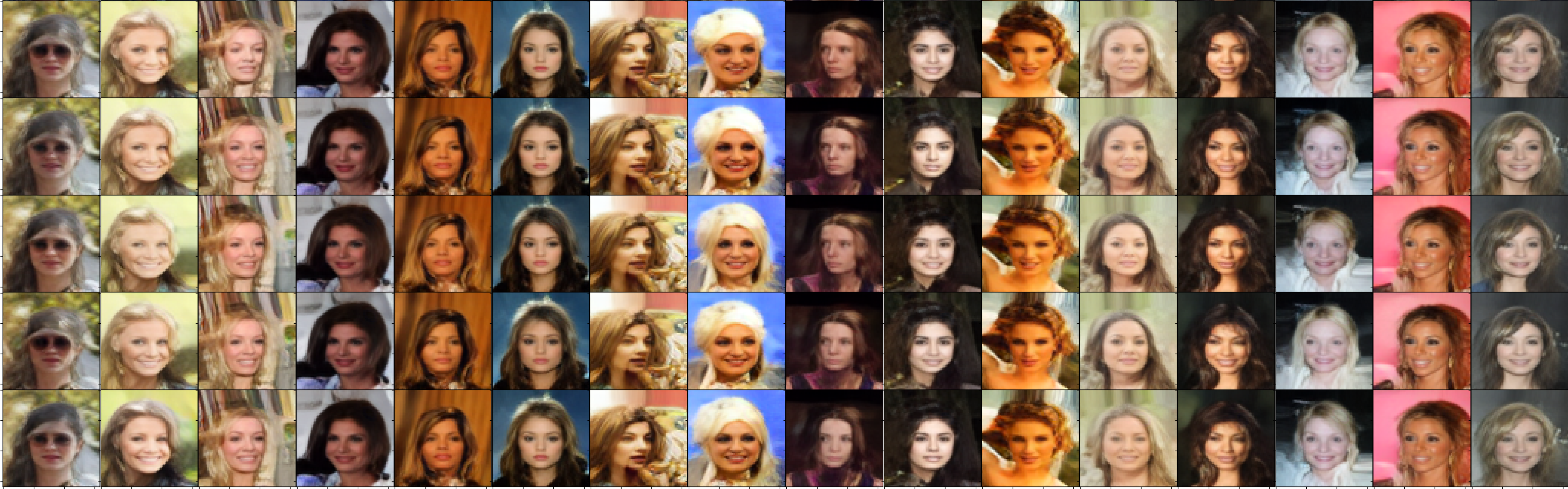}
        \subcaption{Different generated female images by \flowname model}
    \end{minipage}
    \caption{Additional test images of males (in true $P$, figure (a)) and corresponding generated female images by \flowname model (in generated $Q$, figure (b)). To generate different females for a given male, we sample random latent codes given by the VAE, map them through trained \flowname model, and decode back through the VAE decoder to visualize different generated females in the pixel space.}
    \label{fig:celeba_more}
\end{figure}

\subsection{Image-to-image translation}\label{app:handbag_shoes}

The dataset of handbags $P$ has 137K images and the dataset of shoes $Q$ has 50K images, which are (3,64,64) RGB images. The dataset of CelebA males $P$ has 90K images and the dataset of CelebA females $Q$ has 110K images, which are (3,64,64) RBG images as well. Following \citep{korotin2023neural}, we train on 90\% of the total data from $P$ and $Q$ and reserve the rest 10\% data as the test set. The FIDs are computed between generated shoe/female images (from the test handbag/male images) and true shoe/female images from the test set.

We first train a single deep VAE on both $P$ and $Q$. We train the deep VAE in an adversarial manner following \citep{esser2021taming}. Specifically, given a raw image input $X$, the encoder $\mathcal{E}$ of the VAE maps $X$ to $(\mu(X),\Sigma(X))$ parametrizing a multivariate Gaussian of dimension $d$. Then, the VAE is trained so that for a random latent code $X_{enc}\sim \calN(\mu(X),\Sigma(X))$, the decoded image $\mathcal{D}(X_{enc})\approx X$. In our case, each latent code $X_{enc}$ has shape $(12,8,8)$, so that $d=768$. 

The training data for \flowname are thus sets of random latent codes $X_{enc}$ (obtained from $X\sim P$) and $Y_{enc}$ (obtained from $Y\sim P$), where \flowname finds the dynamic OT between the marginal distributions of $X_{enc}$ and $Y_{enc}$. We then obtain the trajectory between $P$ and $Q$ by mapping the OT trajectory in latent space by the decoder $\mathcal{D}$

In Q-flow, the flow architecture $f(x(t),t;\theta)$ consists of convolutional layers of dimensions 12-64-256-512-512-1024, followed by convolutional transpose layers whose filters mirror the convolutional layers. The kernel sizes are 3-3-3-3-3-3-3-4-3-3 with strides 1-1-2-1-1-1-1-2-1-1. We use the softplus activation with $\beta=20$. The initialization of the flow is done by the method of Interflow \citep{albergo2023building}, where at each step, we draw random batches of 128 $X$ and 128 $Y$ and then obtain 128 random latent codes $X_{enc}$ and 128 $Y_{enc}$. We trained the initialized flow for 20K steps. 

To apply Algorithm \ref{alg:refine}, we let $\gamma=0.05$ (bag-shoe) or $\gamma=0.1$ (male-female), $t_k=k/10$ for $k=0,\ldots,10$, and \texttt{Tot}=1. The selection of $\gamma$ is based on grid searching $\gamma$ within 1e-5 and 1 to find one that leads to the highest FID on training data. Specifically, we randomly pick two subsets of training images (with the same size as corresponding test sets) from $P$ and $Q$, obtain the translated images of $P$ images via trained Q-flow net, and compute the FID between translated images and images sampled from $Q$.
Meanwhile, the classifiers are initially trained for 4000 batches with batch size 512 and updated every 10 batches of training $\theta$ for 20 inner-loop batches. The architecture of the classifier networks $c_1$ and $\tilde{c}_0$ is based on \citep{choi2022density}, where the encoding layers of the classifier are convolutional filters of sizes 12-256-512-512-1024-1024 with kernel size equal to 3 and strides equal to 1-1-2-1-1. The decoding layers of the classifier resemble the encoding layers, and the final classification is made by passing the deep decoded feature through a fully connected network with size 768-768-768-1.
Lastly, we train flow parameters $\theta$ for 30K batches (bag-shoe) or for 18K batches (male-female) with a batch size of 256. 

\subsection{High-dimensional density ratio estimation}
\subsubsection{Toy data in 2d}\label{appendix_2d}

\paragraph{Gaussian mixtures.}\label{appendix_2dGMM}

\begin{figure}[!b]
    \centering
    \begin{minipage}{0.19\textwidth}
    \includegraphics[width=\linewidth]{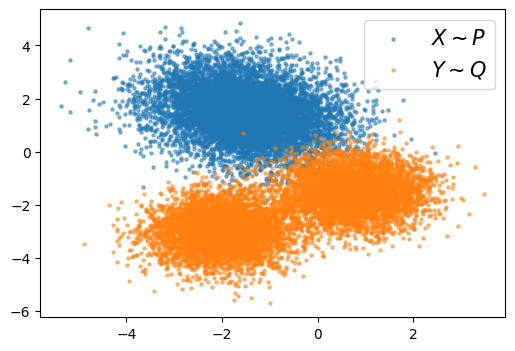}
    \subcaption{Samples}
    \label{2dGMM_data}
    \end{minipage}
    \begin{minipage}{0.19\textwidth}
    \includegraphics[width=\linewidth]{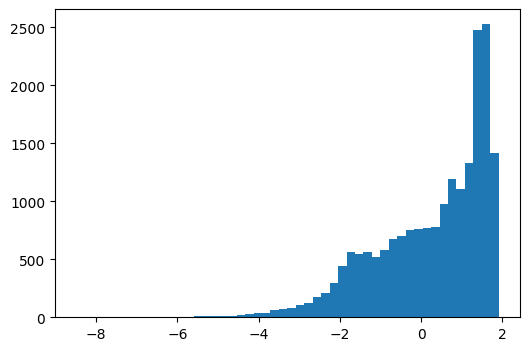}
    \subcaption{MAE,\ours{}:~2.38}
    \end{minipage}
    \begin{minipage}{0.19\textwidth}
    \includegraphics[width=\linewidth]{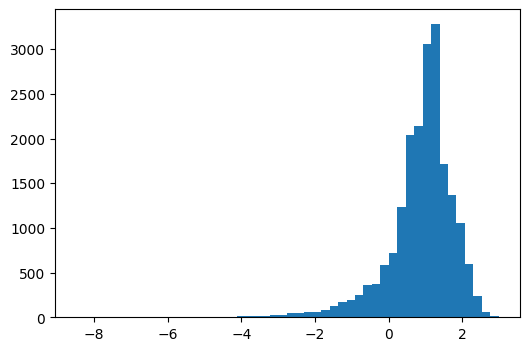}
    \subcaption{\DREinf{}: 3.22}
    \end{minipage}
    \begin{minipage}{0.19\textwidth}
    \includegraphics[width=\linewidth]{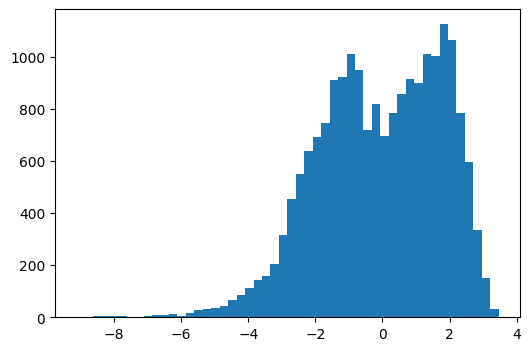}
    \subcaption{TRE: 3.05}
    \end{minipage}
    \begin{minipage}{0.19\textwidth}
    \includegraphics[width=\linewidth]{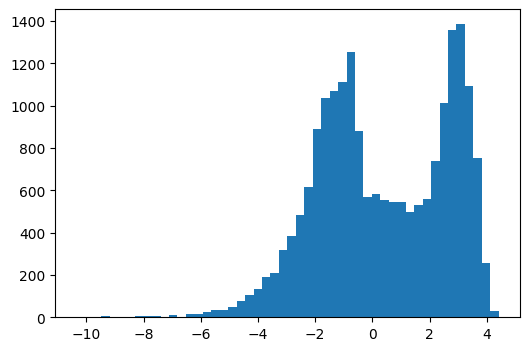}
    \subcaption{1 ratio: 8.20}
    \end{minipage}
    \hspace{0.2in}
    \begin{minipage}{0.99\textwidth}
        \includegraphics[width=\linewidth]{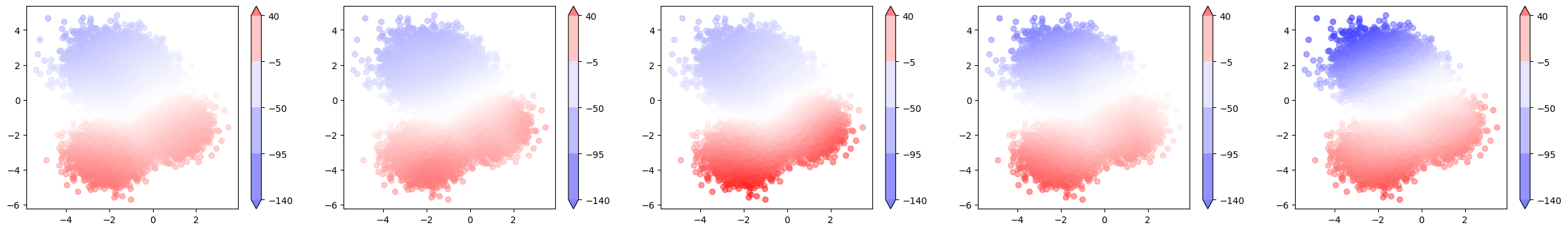}
        \subcaption{Left to right: True ratio $r(x)$, $\hat{r}(x)$ by \ours{}, \DREinf{}, TRE, and 1 ratio.}
    \end{minipage}
    \caption{Estimated log density ratio between 2D Gaussian mixture distributions $P$ (three components) and $Q$ (two components). 
    \textbf{Top:} 
    (a) training samples from $P$ and $Q$. 
    (b)-(d) histograms of errors $\log(|r(x)-\hat{r}(x)|)$ computed at 10K test samples shown in log-scale.
    The MAE \eqref{2dGMM_loss} are shown in the captions. \textbf{Bottom:} true and estimated $\log(q/p)$
    from different models shown under shared color bars.}
    \label{fig:2dGMM}
\end{figure}

\textit{Setup}: We design the Gaussian mixtures $P$ and $Q$ as follows: 
\begin{align*}
    & P=\frac{1}{3}\left(\calN(\begin{bmatrix}
    -2 \\ 2
\end{bmatrix}, 0.75I_2)+\calN(\begin{bmatrix}
    -1.5 \\ 1.5
\end{bmatrix}, 0.25I_2)+\calN(\begin{bmatrix}
    -1 \\ 1
\end{bmatrix}, 0.75I_2) \right) \\
& Q = \frac{1}{2} \left(\calN(\begin{bmatrix}
    0.75 \\ -1.5
\end{bmatrix}, 0.5I_2)+\calN(\begin{bmatrix}
    -2 \\ -3
\end{bmatrix}, 0.5I_2) \right).
\end{align*}
Then, 60K training samples and 10K test samples are randomly drawn from $P$ and $Q$. We intentionally designed the Gaussian mixtures so that their supports barely overlap. The goal is to estimate the $\log$-density ratio $r(x)=\log q(x)-\log p(x)$ on test samples.

Given a trained ratio estimator $\hat{r}(x)$, we measure its performance based on the MAE
\begin{equation}\label{2dGMM_loss}
    \frac{1}{N'} \sum_{i=1}^{N'} |r(X_i)-\hat{r}(X_i)| + \frac{1}{M'} \sum_{j=1}^{M'} |r(\tilde{X}_j)-\hat{r}(\tilde{X}_j)|,
\end{equation}
where we use $N'$ and $M'$ test samples from $P$ and $Q$, and $r(x)$ denotes the true density between $P$ and $Q$.

\noindent \textit{\flowname}: To initialize the two \JKO{} models that consists of the initial \flowname, we specify the \JKO{} as:
\begin{itemize}
    \item The flow network $f(x(t),t;\theta_P)$ and $f(x(t),t;\theta_Q)$ consists of fully-connected layers 
    
    \texttt{3$\rightarrow$128$\rightarrow$128$\rightarrow$2}. The Softplus activation with $\beta=20$ is used. We concatenate $t$ along $x$ to form an augmented input into the network.
    
    \item We train the initial flow with a batch size of 2000 for 100 epochs along the grid 
    
    $[0,0.25), [0.25,0.625), [0.625, 1)$.
\end{itemize}
\begin{table}[!b]
    \centering
    \caption{
    Inversion error 
    $\mathbb{E}_{x\sim P }\|{T_1^0}({T_0^1}(x))-x\|^2_2+\mathbb{E}_{y\sim Q }\|{T_0^1}({T_1^0}(y))-y\|^2_2$ of \flowname
    computed via sample average on the test split of the data set.
    }\label{inv_err}
    \renewcommand{\arraystretch}{1.5}
    \resizebox{0.75\linewidth}{!}{
    \begin{tabular}{c|c|c}
        \toprule
         moon-to-checkerboard & High-dimensional Gaussians ($d=320$) & MNIST $(Q $ by RQ-NSF)  \\
       \hline
         7.24e-7 & 3.44e-5 & 5.23e-5\\
         \bottomrule
    \end{tabular}}
\end{table}
To refine the \flowname, we concatenate the trained $f(x(t),t;\theta_P)$ and $f(x(t),t;\theta_Q)$, where the former flows in $[0,1)$ to transport $P$ to $Z$ and the latter flows in $[1,0)$ to transport $Z$ to $Q$. We then use the time grid $[0,0.25), [0.25,0.625), [0.625, 1),[1,0.625),[0.625, 0.25),[0.25,0)$ to train $f(x(t),t;\theta)$ with $\theta = \{\theta_P, \theta_Q\}$; we note that the above time grid can be re-scaled to obtain the time grid $\{t_k\}_{k=1}^K$ over $[0,1]$. The hyperparameters for Algorithm \ref{alg:refine} are: \texttt{Tot}=2, $E_0=300$, $E=50$, $E_{\rm in}=4, \gamma = 0.5$. The classification networks $\{c_1,\widetilde{c}_0\}$ consists of fully-connected layers \texttt{2$\rightarrow$312$\rightarrow$312$\rightarrow$312$\rightarrow$1} with the Softplus activation with $\beta=20$, and it is trained with a batch of 200.

\noindent \textit{flow-ratio}: The network consists of fully-connected layers \texttt{3$\rightarrow$256$\rightarrow$256$\rightarrow$256$\rightarrow$1} with the Softplus activation with $\beta=20$. The input dimension is 3 because we concatenate time $t$ along the input $x\in \R^2$ to form an augmented input. Using the trained \flowname model, we then produce a bridge of 6 intermediate distributions using the pre-scaled grid $[0,0.25), [0.25,0.625), [0.625, 1),$ $[1,0.625),[0.625, 0.25),[0.25,0)$ for the \flowname. We then train the network $r(x,t;\theta_r)$ for 100 epochs with a batch size of 1000, corresponding to \texttt{Tot\_iter}=6K in Algorithm \ref{alg:rnet}.

\begin{figure}[!t]
    \centering
    \includegraphics[width=0.75\linewidth]{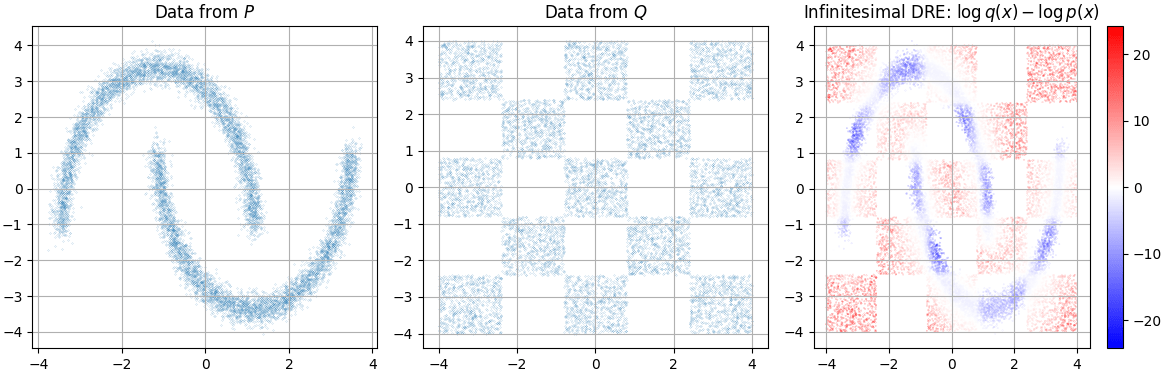}
    \caption{The \rationame net estimation of $\log (q(x)/p(x))$ (right figure), evaluated at the union of test data from $P$ and $Q$ (left and middle figures).}
    \label{moon_to_checkerboard_PQ}
\end{figure}
\paragraph{Two-moon to and from checkerboard.}\label{appendix_moon_to_checkerboard}
\textit{Setup}: We generate 2D samples whose marginal distribution has the shape of two moons and a checkerboard (see Figure \ref{moon_to_checkerboard}(a), leftmost and rightmost scatter plots). We randomly sample 100K samples from $P$ and $Q$ to train the \flowname and the infinitesimal DRE.

\noindent \textit{\flowname}: To initialize the two \JKO{} models that consists of the initial \flowname, we specify the \JKO{} as:
\begin{itemize}
    \item The flow network $f(x(t),t;\theta_P)$ and $f(x(t),t;\theta_Q)$ consists of fully-connected layers 
    
    \texttt{3$\rightarrow$256$\rightarrow$256$\rightarrow$2}. The Softplus activation with $\beta=20$ is used. We concatenate $t$ along $x$ to form an augmented input into the network.

    \item We train the initial flow with a batch size of 2000 for 100 epochs along the grid  
    
    $[0,0.25), [0.25,0.5), [0.5,0.75),$ $[0.75, 1)$.
\end{itemize}

\begin{wrapfigure}[12]{r}{0.6\textwidth}
\centering
\vspace{-15pt}
\includegraphics[width = 0.95\linewidth]{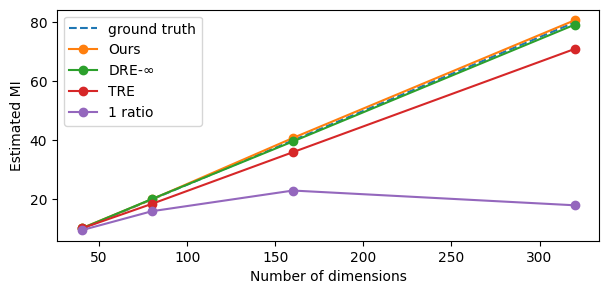}
\vspace{-5pt}
\caption{Estimated MI between two correlated high-dimensional Gaussian random variables.}
\label{fig:MI_gaussian}
\end{wrapfigure}
To refine the \flowname, we concatenate the trained $f(x(t),t;\theta_P)$ and $f(x(t),t;\theta_Q)$, where the former flows in $[0,1]$ to transport $P$ to $Z$ and the latter flows in $[1,0]$ to transport $Z$ to $Q$. We then use the grid $[0,0.25), [0.25,0.5),$ $[0.5,0.75), [0.75, 1), [1,0.75),[0.75, 0.5),$ $[0.5, 0.25),[0.25,0)$ (which can be re-scaled to form the time grid over $[0,1)$) to train $f(x(t),t;\theta)$ with $\theta = \{\theta_P, \theta_Q\}$.
The hyperparameters for Algorithm \ref{alg:refine} are: \texttt{Tot}=2, $E_0=300$, $E=50$, $E_{\rm in}=4, \gamma = 0.5$. The classification networks $\{c_1,\widetilde{c}_0\}$ consists of fully-connected layers \texttt{2$\rightarrow$312$\rightarrow$312$\rightarrow$312$\rightarrow$1} with the Softplus activation with $\beta=20$, and it is trained with a batch of 200.

\noindent \textit{flow-ratio}: The network consists of fully-connected layers \texttt{3$\rightarrow$256$\rightarrow$256$\rightarrow$256$\rightarrow$1} with the Softplus activation with $\beta=20$. The input dimension is three because we concatenate time $t$ along the input $x\in \R^2$ to form an augmented input. Using the trained \flowname model, we then produce a bridge of 8 intermediate distributions using the pre-scaled grid interval $[0,0.25), [0.25,0.5), [0.5,0.75)$, $[0.75, 1), [1,0.75),$ $[0.75, 0.5), [0.5, 0.25),[0.25,0)$ for the \flowname. We then train the network $r(x,t;\theta_r)$ for 500 epochs with a batch size of 500, corresponding to \texttt{Tot\_iter}=100K in Algorithm \ref{alg:rnet}.

\subsubsection{High-dimensional Mutual Information estimation}\label{appendix_MI}

\textit{Setup}: 
We follow the same setup as in \citep{rhodes2020telescoping,choi2022density}.
The first Gaussian distribution $P=\calN(0,\Sigma)$, where $\Sigma$ is a block-diagonal covariance matrix with $2\times 2$ small blocks having one on the diagonal and 0.8 on the off-diagonal. The second Gaussian distribution $Q=\calN(0,I_d)$ is the isotropic Gaussian in $\R^d$. We randomly draw 100K samples for each choice of $d$, which varies from 40 to 320.

To be more precise, we hereby draw the connection of the DRE task with mutual information (MI) estimation, following \citep{rhodes2020telescoping}. We first recall the definition of MI between two correlated random variables $U$ and $V$:
\begin{equation}\label{eq:MI}
    I(U;V)=\mathbb{E}_{p(U,V)}\left[\log \frac{p(U,V)}{p(U)p(V)}\right].
\end{equation}
Now, given $X=(x_1,\ldots,x_d)\sim P= \calN(0,\Sigma)$, we define $U=(x_1,x_3,\ldots,x_{d-1})$ and $V=(x_2,x_4,\ldots,x_d)$. By the construction of $\Sigma$, we thus have $p(U)p(V)=Q(X)$ for $Q=\calN(0,I_d)$. As a result, the MI in \eqref{eq:MI} between $U$ and $V$ is equivalent to $\mathbb{E}_{X\sim P}[-r(X)]$, where $r(x)=\log \frac{Q(x)}{P(x)}$ is the objective of interest in DRE.

\noindent \textit{\flowname}: 
We specify the following when training the \flowname:
\begin{itemize}
    \item The flow network $f(x(t),t;\theta)$ consists of fully-connected layers with dimensions 
    
    \texttt{(d+1)$\rightarrow$min(4d,1024)$\rightarrow$min(4d,1024)$\rightarrow$d}. The Softplus activation with $\beta=20$ is used. We concatenate $t$ along $x$ to form an augmented input into each network layer.

    \item We train the flow network for 100 epochs with a batch size of 500, in both the flow initialization phase and the end-to-end refinement phase.
    The flow network is trained along the evenly-spaced time grid $[t_{k-1},t_k)$ for $k=1,\ldots,L_d$, and we let $t_k=k/L_d$. 
    $L_d$ increases as the dimension $d$ increases.
    We specify the choices as 
    
    $(L_d,d) \in \{(4,40), (6,80), (7,160), (8, 320)\}$.
\end{itemize}

The hyperparameters for Algorithm \ref{alg:refine} are: \texttt{Tot}=2, $E_0=500$, $E=100$, $E_{\rm in}=2, \gamma = 0.5$. The classification networks $\{c_1,\widetilde{c}_0\}$ consists of fully-connected layers \texttt{$d\rightarrow \min(4d,1024)\rightarrow\min(4d,1024)$}\texttt{$\rightarrow\min(4d,1024)\rightarrow 1$} with the Softplus activation with $\beta=20$, and it is trained with a batch of 200.

\noindent \textit{flow-ratio}: The network consists of fully-connected layers with dimensions 

\noindent \texttt{(d+1)$\rightarrow$min(4d,1024)$\rightarrow$min(4d,1024)$\rightarrow$min(4d,1024)$\rightarrow$1}, using the Softplus activation with $\beta=20$. The input dimension is $d+1$ because we concatenate time $t$ along the input $x\in \R^d$ to form an augmented input.
Using the trained \flowname model, we then produce a bridge of $L_d$ intermediate distributions using the grid $[t_{k-1},t_k)$ specified above for the \flowname. We then train the network $r(x,t;\theta_r)$ for 1000 epochs with a batch size of 512, corresponding to \texttt{Tot\_iter}=195K in Algorithm \ref{alg:rnet}.

\begin{figure}[!b]
   \centering
    \includegraphics[width=\textwidth]{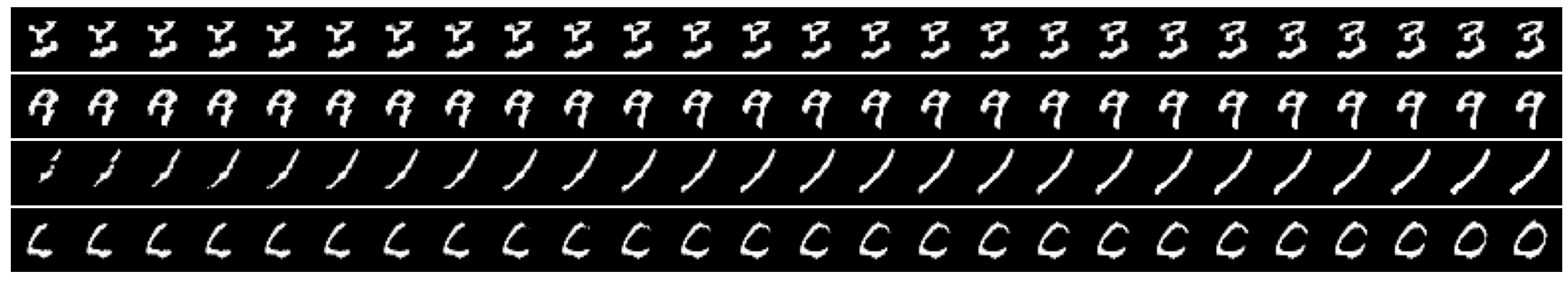}
    \caption{The trajectory of samples (in rows) from intermediate distributions of the Q-flow, as it pushes forward the base distribution (leftmost column) to the target distribution (rightmost column). The figure shows the improvement of generated digits using the Q-flow.}
    \label{fig:RQ_NSF_improved_traj}
\end{figure}
\subsubsection{Energy-based modeling of MNIST}\label{appendix_MNIST}

\begin{table}[!t]
    \centering
    \caption{DRE performance on the energy-based modeling task for MNIST, reported in BPD and lower, is better. We train \rationame net using trajectories given by bi-directional and uni-directional Q-flow. The uni-directional Q-flow only optimizes \eqref{tigheMtoL} along the $P\rightarrow Q$ direction.}
    \label{tab:bpd_bidir}
    \renewcommand{\arraystretch}{1}
    \setstretch{1.25}
    {\fontsize{30pt}{30pt}\selectfont
    \resizebox{0.7\linewidth}{!}{
    \begin{tabular}{c|ccc}
    \toprule
 & RQ-NSF & Copula & Gaussian   \\
\hline
\rationame net using bi-directional \flowname & 1.05 & 1.14 & 1.31\\
\hline
\rationame net using uni-directional \flowname & 1.08 & 1.19 & 1.31\\
\bottomrule
    \end{tabular}
    }}
\end{table}

\begin{figure}[!t]

    \begin{minipage}{0.49\textwidth}
    \includegraphics[width=\linewidth]{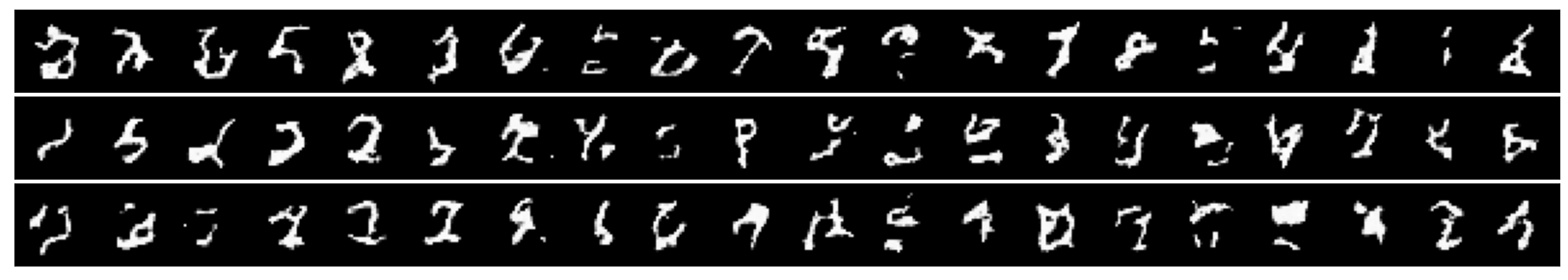}
    \subcaption{RQ-NSF: raw samples from $Q$}
    \end{minipage}
    \begin{minipage}{0.49\textwidth}
    \includegraphics[width=\linewidth]{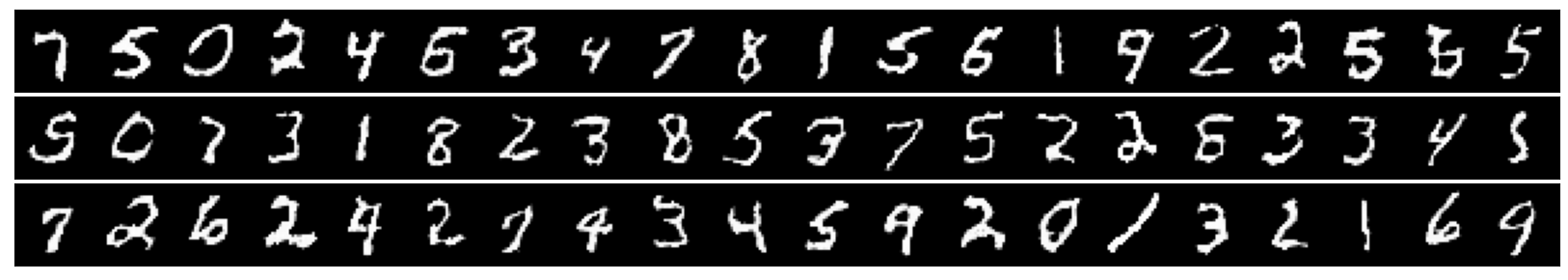}
    \subcaption{RQ-NSF: improved samples from $\tilde{Q}$}
    \end{minipage}

    \begin{minipage}{0.49\textwidth}
    \includegraphics[width=\linewidth]{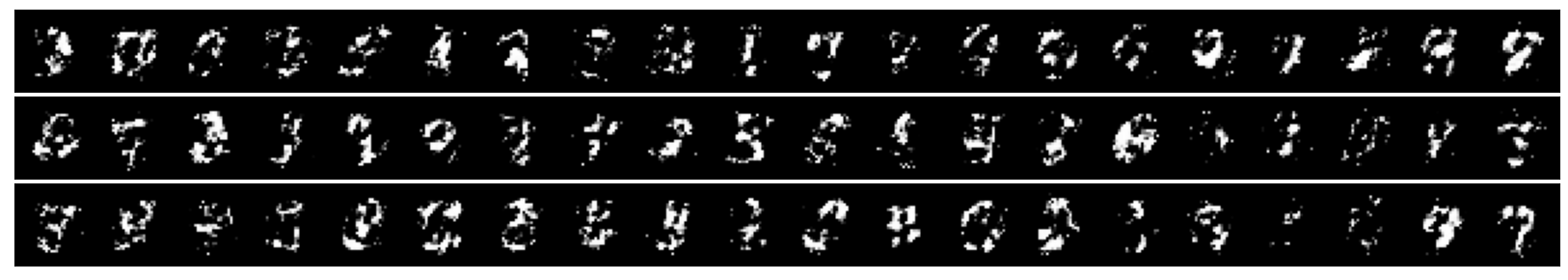}
    \subcaption{Copula: raw samples from $Q$}
    \end{minipage}
    \begin{minipage}{0.49\textwidth}
    \includegraphics[width=\linewidth]{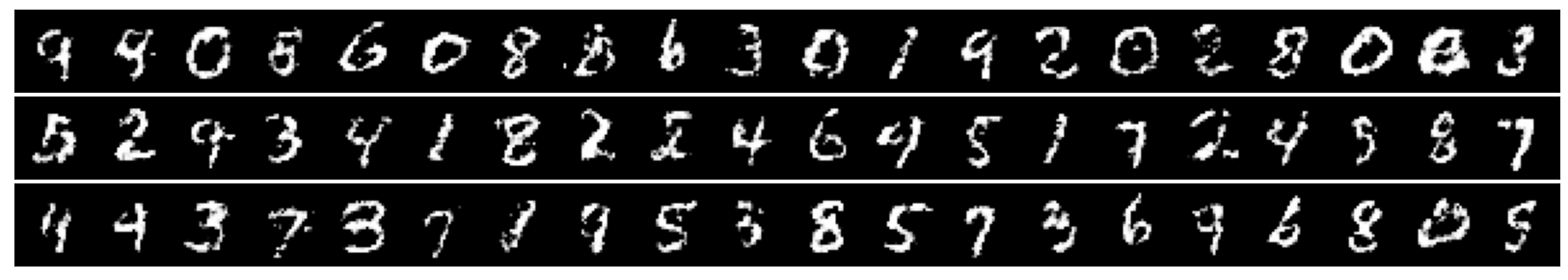}
    \subcaption{Copula: improved samples from $\tilde{Q}$}
    \end{minipage}

    \begin{minipage}{0.49\textwidth}
    \includegraphics[width=\linewidth]{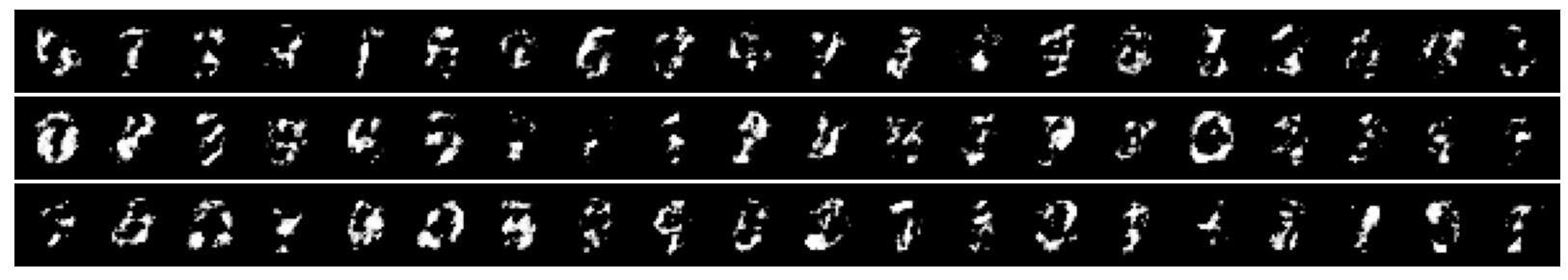}
    \subcaption{Gaussian: raw samples from $Q$}
    \end{minipage}
    \begin{minipage}{0.49\textwidth}
    \includegraphics[width=\linewidth]{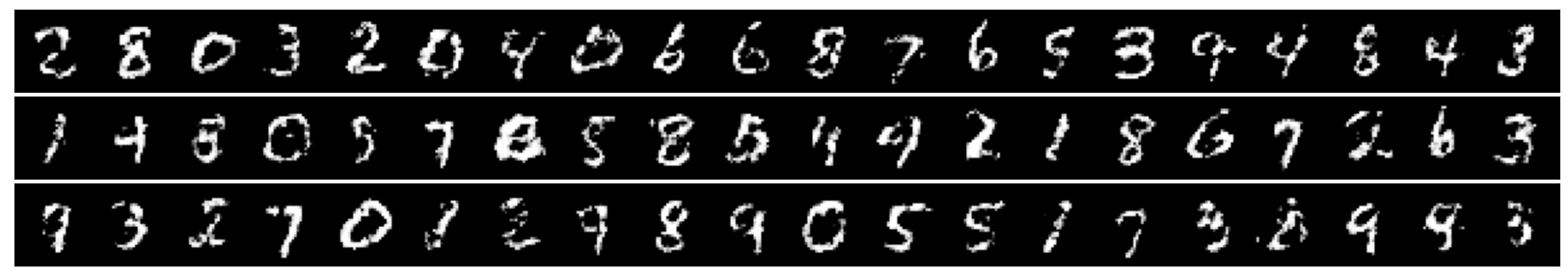}
    \subcaption{Gaussian: improved samples from $\tilde{Q}$}
    \end{minipage}
    
    \caption{Improvement in generated samples from $Q$, where $Q$ is given by RQ-NSF, Copula, or Gaussian. Each of the three choices of $Q$ is defined by a pre-trained invertible model $F$ that yields $Q=F_{\#}\calN(0,I_d)$.}
    \label{fig:MNIST_improved}
\end{figure}

\noindent \textit{Setup.} 
We discuss how each of the three $Q$ distributions is obtained based on \citep{rhodes2020telescoping,choi2022density} and how we apply \flowname and \rationame nets to the problem.
Specifically, the MNIST images are in dimension $d=28^2=784$, and each of the pre-trained models provides an invertible mapping $F: \R^d \to \R^d$, where $Q = F_\# \calN(0,I_d)$.
We train a \flowname net between $(F^{-1})_{\#} P$ and $(F^{-1})_{\#} Q$, the latter by construction equals $\calN(0,I_{d})$.
Using the trained \flowname net, we go back to the input space and train the \rationame net using the intermediate distributions between $P$ and $Q$.

The trained \rationame $r(x,s;\hat{\theta}_r)$ provides an estimate of the data density $p(x)$ by $\hat{p}(x)$ defined as 
$\log \hat{p}(x) = \log q(x) - \int_0^1 r(x,s;\hat{\theta}_r)ds$,
where $\log q(x)$ is given by the change-of-variable formula using the pre-trained model $F$ and the analytic expression of $\calN(0,I_d)$.
As a by-product, since our \flowname net provides an invertible mapping $T_0^1$, we can use it to obtain an improved generative model on top of $F$. Specifically, 
the improved distribution $\tilde{Q}: =(F \circ T^0_1)_{\#} \calN(0,I_{d})$, that is,  we first use \flowname to transport $\calN(0,I_{d})$ and then apply $F$.
The performance of the improved generative model can be measured using the ``bits per dimension'' (BPD) metric: 
\begin{equation}\label{eq:bpd}
\text{BPD}= \frac{1}{N'} \sum_{i=1}^{N'} [- \log \hat{p}(X_i)/ (d \log 2)],
\end{equation}
where $X_i$ are $N'$=10K test images drawn from $P$. 
BPD has been a widely used metric in evaluating the performance of generative models \citep{theis2016note,papamakarios2017masked}.
In our setting, the BPD can also be used to compare the performance of the DRE.

\noindent \textit{\flowname}: 
We specify the following when training the \flowname:
\begin{itemize}
    \item The flow network $f(x(t),t;\theta)$ consists of fully-connected layers with dimensions 
    
    \texttt{(d+1)$\rightarrow$1024$\rightarrow$1024$\rightarrow$1024$\rightarrow$d}. The Softplus activation with $\beta=20$ is used. We concatenate $t$ along $x$ to form an augmented input into the network.

    \item In a block-wise fashion, we train the network with a batch size of 1000 for 100 epochs along the grid $[t_{k-1}, t_{k-1}+h_k)$ for $k=1,\ldots,5$. We let $h_k=0.5\cdot1.1^{k-1}$. 
\end{itemize}
The hyperparameters for Algorithm \ref{alg:refine} are: \texttt{Tot}=2, $E_0=100$, $E=500$, $E_{\rm in}=2, \gamma = 0.5$. The classification networks $\{c_1,\widetilde{c}_0\}$ consists of fully-connected layers \texttt{$784 \rightarrow 1024 \rightarrow 1024$}\texttt{$\rightarrow 1024 \rightarrow$1} with the Softplus activation with $\beta=20$, and it is trained with a batch of 200.

\noindent \textit{flow-ratio}: We use the same convolutional U-Net as described in \citep[Table 2]{choi2022density}, which consists of an encoding block and a decoding block comprised of convolutional layers with varying filter sizes. Using the trained \flowname model, we then produce a bridge of 5 intermediate distributions using the intervals $[t_{k-1}, t_{k-1}+h_k)$ specified above for the \flowname. We then train the network $r(x,t;\theta_r)$ for 300 epochs with a batch size of 128, corresponding to \texttt{Tot\_iter}=117K in Algorithm \ref{alg:rnet}.

\subsection{Hyper-parameter sensitivity}\label{sec:hyper_params}
\begin{wraptable}[10]{r}{0.5\textwidth}
    \vspace{-15pt}
    \centering
    \caption{BPD on MNIST with RQ-NSF target $Q$ over combinations of $\gamma$ in Algorithm \ref{alg:refine} and time grid $\{t_k\}$ in Algorithm \ref{alg:rnet}.}\label{tab:ablation}
    \renewcommand{\arraystretch}{1.5}
    \resizebox{\linewidth}{!}{
    \begin{tabular}{c|c|c|c}
        \toprule
        $\gamma \ \& \ \{t_k\}$ & $t_k=k/L$ & $t_k=(k/L)^2$ & $t_k=\sqrt{k/L}$  \\
       \hline
        0.5 & 1.046 & 1.044 & 1.047 \\
        1 & 1.042 & 1.041 & 1.044  \\
        5 & 1.057 & 1.055 & 1.062  \\
         \bottomrule
    \end{tabular}}
\end{wraptable}

Overall, we did not purposely tune the hyperparameters in Section \ref{sec_experiments}, and found that the Algorithm \ref{alg:refine} and \ref{alg:rnet} are not sensitive to hyper-parameter selections. We conduct additional ablation studies by varying the combination of $\gamma$ in Algorithm \ref{alg:refine} and the time grid $\{t_k\}$ in Algorithm \ref{alg:rnet}. We tested all combinations on the MNIST example in Section \ref{expr:mnist} with RQ-NSF target $Q$. Table \ref{tab:ablation} below presents our method's performance, with the highest BPD (1.062) remaining lower than those by other DRE baselines in Table \ref{tab:bpd} (the lowest of which is 1.09). Small variations in the table can be attributed to the learned OT trajectory influenced by the choice of $\gamma$. Specifically, smaller $\gamma$ may lead to less smooth trajectories between $P$ and $Q$. In contrast, larger $\gamma$ may result in a higher KL-divergence between the pushed and target distributions due to insufficient amount of distribution transportation by the refined flow, both potentially impacting DRE accuracy.

\begin{figure}[!t]
    \centering
    \includegraphics[width=\linewidth]{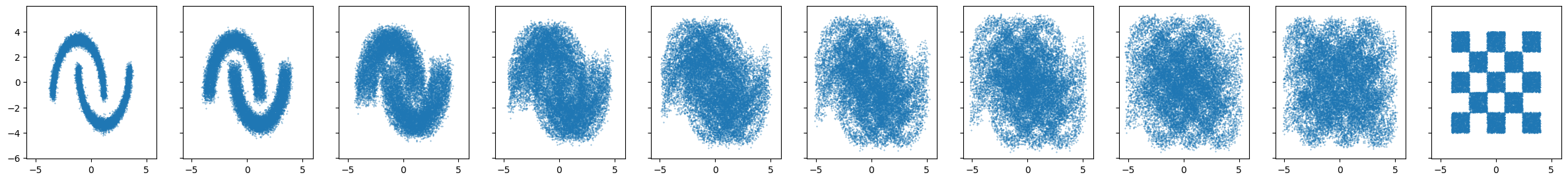}
    \caption{Bridge construction between $P$ (leftmost) and $Q$ (rightmost) via the linear interpolation scheme \eqref{eq:linear_interpolate}. Specifically, we choose $\alpha_k = k/9$ for $k=0,\ldots,9$.}
\label{fig:linear_moon_to_checkerboard_interpolate}
\end{figure}

\section{Additional methodology details}

\subsection{Computational complexity of end-to-end training}\label{app:comp-complexity}

We measure the computational complexity by the number of function evaluations of $f(x(t),t;\theta)$ and of the classification nets $\{c_1, \widetilde{c}_0\}$. 
Suppose the total number of epochs in outer loop training is $O(E)$; the dominating computational cost lies in the neural ODE integration, which takes ${O}(8 KS \cdot E(M+N) )$ function evaluations of $f(x,t;\theta)$.
We remark that the Wasserstein-2 {loss} \eqref{eq:min_W2} 
incurs no extra computation,
since the samples $X_i(t_k;\theta)$ and $\tilde{X}_j(t_k;\theta)$ are available when computing the forward and reverse time integration of $f(x,t; \theta)$.
The training of the two classification nets $c_1$ and $\widetilde{c}_0$  takes ${O}(4 (E_0+  E E_{\rm in})(M+N) )$ additional evaluations of the two network functions since the samples $X_i(1;\theta)$ and $\tilde{X}_j(0;\theta)$ are already computed.

\subsection{Details of flow initialization}\label{app:flow-init}

We consider two initialization schemes:

(i) By a concatenation of two CNF models.

Each of the two CNF models flows invertibly between $P$ and $Z$ and $Z$ and $Q$ respectively, where $Z\sim \calN(0, I_d)$. Any existing neural-ODE CNF models may be adopted for this initialization \citep{FFJORD,xu2023normalizing}.

(ii) By distribution interpolant neural networks.

Specifically, one can use the linear interpolant mapping in \citep{rhodes2020telescoping,choi2022density,albergo2023building} as below, and train the neural network velocity field $f(x,t;\theta)$ to match the interpolation \citep{albergo2023building}.

The interpolation scheme used in \citep[Eq (5)]{rhodes2020telescoping} states that given a pair of random samples $X(0) \sim P$ and $X(1) \sim Q$, the interpolated sample $X(t_k)$ is defined as
\begin{equation}\label{eq:linear_interpolate}
    X(t_k) = \sqrt{1-\alpha_k^2} X(0) + \alpha_k X(1),
\end{equation}
where $\alpha_k$ forms an increasing sequence from 0 to 1.
An illustration of $\{X(t_k)\}$ is given in Figure \ref{fig:linear_moon_to_checkerboard_interpolate}.

\end{document}